\RequirePackage
[
abort,
l2tabu,
orthodox,
]
{nag}\RequirePackage{import}\makeatletter
\def\ece#1#2{\expandafter#1\csname#2\endcsname}\def\setproperty#1#2#3{\ece\protected@edef{#1@p#2}{\unexpanded{#3}}}
\def\getproperty#1#2{\expandafter\ifx\csname#1@p#2\endcsname\relax \else \csname#1@p#2\endcsname \fi }\def\ifthenelsepropertydefined#1#2#3#4{\expandafter\ifx\csname#1@p#2\endcsname\relax #4 \else#3 \fi }\def\ifpropertydefined#1#2#3{\ifthenelsepropertydefined{#1}{#2}{#3}{}}
\def\ifpropertyundefined#1#2#3{\ifthenelsepropertydefined{#1}{#2}{}{#3}}
\def\raiseifpropertyundefined#1#2#3{\ifpropertyundefined{#2}{#3}{\PackageError{#1}{Property #2 #3 needs to be defined. Put \@backslashchar setproperty{#2}{#3} to your settings file}{Grep for your property :)}}
}
\def\setpropertyifundefined#1#2#3{\ifpropertyundefined{#1}{#2}{\setproperty{#1}{#2}{#3}}{}}
\def\setpropertyifundefinedwithusageinfo#1#2#3{\setpropertyifundefined{#1}{#2}{TODO>> Put \backslash{}setproperty\{#1\}\{#2\}\{<your value here,e.g. ``#3``>\} into content/settings.tex <<}
}

\def\ifthenelseproperty#1#2#3#4{\providetoggle{#1@p#2}\settoggle{#1@p#2}{\getproperty{#1}{#2}}\iftoggle{#1@p#2}{#3}{#4}}
\def\ifthenelsepropertyequal#1#2#3#4#5{\ifthenelse{\equal{\getproperty{#1}{#2}}{#3}}{#4}{#5}}

\makeatother
\makeatletter
\newcommand\RequirePackageWithOption[2][]{\@ifpackageloaded{#2}{\PassOptionsToPackage{#1}{#2}}{\RequirePackage[#1]{#2}}
}
\newcommand\ImportDefault[2][]{\IfFileExists{content/#1#2.tex}{\typeout{----- LOAD OWN #1#2 -----}\import{content/#1}{#2}}{\typeout{----- LOAD DEFAULT #1#2 -----}\ifthenelsepropertydefined{default}{#2}{\IfFileExists{templates/\getproperty{default}{#2}/#1#2.tex}{\import{templates/\getproperty{default}{#2}/#1}{#2}}{\import{templates/Default/#1}{#2}}
}{\typeout{\getproperty{default}{#1}}
\ifthenelsepropertydefined{default}{#1}{\IfFileExists{templates/\getproperty{default}{#1}/#1#2.tex}{\import{templates/\getproperty{default}{#1}/#1}{#2}}{\import{templates/Default/#1}{#2}}
}{\import{templates/Default/#1}{#2}}
}
}
}

\makeatother
\documentclass
[
a4paper,
DIV=calc,
BCOR=8mm,
headinclude,
bibliography=totoc,
listof=totoc,
index=totoc,
british,
twoside,
12pt,
openright,
captions=tableheading,
chapterprefix,
twocolumn,
]
{scrartcl}\newif\ifKOMA
\KOMAtrue
\setproperty{default}{document}{Paper}\setproperty{default}{titlepage}{Paper}\setproperty{default}{appendix}{Paper}\setproperty{default}{content}{Paper}\setproperty{compilation}{acknowledgement}{true}\setproperty{compilation}{appendix}{true}\setproperty{compilation}{lof}{false}\setproperty{compilation}{lot}{false}\setproperty{compilation}{toc}{false}\setproperty{compilation}{complementglyphs}{false}
\newcommand{\fvolnorm}{\langle \mathbf{F} \rangle_\mathrm{vol,norm}}
\newcommand{\phifmn}{\varphi_{\mathrm{F},mn}}
\usepackage{authblk}\usepackage{float}\usepackage{fancyhdr}\usepackage{amsmath}\usepackage{amssymb}\usepackage{amsfonts}\usepackage{textcomp}

\usepackage{cuted}\usepackage{calc}\usepackage{needspace}\newlength{\widelinewidth}
\usepackage
[
ngerman,
main=british
]
{babel}\RequirePackageWithOption
[
log-declarations=false
]
{xparse}\usepackage{xpatch}\usepackage{shellesc}\usepackage{ifluatex}\ifluatex

\newcommand{\FirstWord}[1]{\luaexec{tex.print(#1)}}

\else
\fi
\setpropertyifundefined{color}{main}{black}\setpropertyifundefined{color}{secondary}{black}\setpropertyifundefinedwithusageinfo{author}{firstname}{J.R.R.}\setpropertyifundefinedwithusageinfo{author}{familyname}{Tolkien}\setpropertyifundefinedwithusageinfo{author}{acadtitle}{Prof.~Dr.~h.\,c.}\setpropertyifundefinedwithusageinfo{author}{addressstreet}{Way of no return 77}\setpropertyifundefinedwithusageinfo{author}{addresscity}{12345 Hobbingen}\AtBeginDocument{\setpropertyifundefined{author}{address}{\getproperty{author}{addressstreet}\\\getproperty{author}{addresscity}}
}
\setpropertyifundefinedwithusageinfo{author}{dateofbirth}{01.01.1970}\setpropertyifundefinedwithusageinfo{author}{placeofbirth}{Helms Klamm,Rohan}\setpropertyifundefinedwithusageinfo{author}{nationality}{german}\setpropertyifundefinedwithusageinfo{author}{mobile}{+47 173 123456}\setpropertyifundefinedwithusageinfo{author}{phone}{+53 001 123456}\setpropertyifundefinedwithusageinfo{author}{faxnr}{+53 001 123456}\setpropertyifundefinedwithusageinfo{author}{email}{minna@barnhelm.com}\setpropertyifundefinedwithusageinfo{author}{orcid}{1234567}\setpropertyifundefinedwithusageinfo{author}{signature}{\backslash{}igraph\{path to your signature here>\}}
\setpropertyifundefined{author}{affiliationindices}{1}\setpropertyifundefined{affiliations}{1}{Max-Planck-Institute for Plasma Physics,17491 Greifswald,Germany}\setpropertyifundefined{document}{date}{\today}\setpropertyifundefinedwithusageinfo{document}{title}{The Lord of the Rings}\setpropertyifundefinedwithusageinfo{document}{titlehead}{There and back again}\setpropertyifundefinedwithusageinfo{document}{subject}{Novel}\setpropertyifundefinedwithusageinfo{document}{keywords}{ring,mordor}\setpropertyifundefinedwithusageinfo{document}{location}{Rivendale}\setpropertyifundefinedwithusageinfo{document}{publishers}{Elrond Verlag}\setpropertyifundefinedwithusageinfo{document}{dedication}{For my precious}\setpropertyifundefinedwithusageinfo{document}{logos}{\backslash{}igraph\{<path to your logo here>\}}
\setpropertyifundefinedwithusageinfo{document}{type}{book}\setpropertyifundefinedwithusageinfo{document}{doi}{145125.1234150}\setpropertyifundefined{compilation}{externalfolder}{tikz/}\setpropertyifundefined{compilation}{externalize}{false}\setpropertyifundefined{compilation}{pdfa}{false}\setpropertyifundefined{compilation}{final}{false}\setpropertyifundefined{compilation}{comment}{footnote}\AtEndPreamble{\ifthenelseproperty{compilation}{pdfa}{\setpropertyifundefined{compilation}{externalize}{false}}{}}
\setpropertyifundefined{compilation}{minionfonts}{false}\setpropertyifundefined{compilation}{libertinusfonts}{false}\setpropertyifundefined{compilation}{complementglyphs}{true}\setpropertyifundefined{compilation}{fancychapterheadings}{false}\setpropertyifundefined{compilation}{microtype}{false}\setpropertyifundefined{compilation}{fontspec}{false}\setpropertyifundefinedwithusageinfo{letter}{recipientname}{Sauron Mustermann}\setpropertyifundefinedwithusageinfo{letter}{recipientaddress}{Ring Institute\\Tower 1\\12345 Mordor}\setpropertyifundefinedwithusageinfo{letter}{subject}{Application for free Ork position}\setpropertyifundefinedwithusageinfo{letter}{opening}{Dear Madam or Sir,}\setpropertyifundefined{letter}{closing}{Sincerely yours,}\setpropertyifundefined{letter}{enclosuretitle}{Attached}\setpropertyifundefined{letter}{enclosure}{curriculum vit\ae{},certificates,references}
\setpropertyifundefined{thesis}{faculty}{FACULTY}\setpropertyifundefined{thesis}{degree}{DEGREE}\setpropertyifundefined{thesis}{defensedate}{DEFENSEDATE}\setpropertyifundefined{thesis}{submissionyear}{YEAR OF SUBMISSION}\makeatletter
\@ifundefined{ifKOMA}{\newif\ifKOMA \KOMAfalse }{}\@ifclassloaded{book}{\setpropertyifundefined{compilation}{titlepage}{true}\setpropertyifundefined{compilation}{abstract}{true}\setpropertyifundefined{compilation}{toc}{true}\setpropertyifundefined{compilation}{bibliography}{true}\setpropertyifundefined{compilation}{lof}{true}\setpropertyifundefined{compilation}{lot}{true}\setpropertyifundefined{compilation}{appendix}{true}\setpropertyifundefined{compilation}{affidavit}{true}\setpropertyifundefined{compilation}{clsdefineschapter}{true}}{}\@ifclassloaded{report}{\setpropertyifundefined{compilation}{titlepage}{true}\setpropertyifundefined{compilation}{abstract}{true}\setpropertyifundefined{compilation}{toc}{true}\setpropertyifundefined{compilation}{bibliography}{true}\setpropertyifundefined{compilation}{lof}{true}\setpropertyifundefined{compilation}{lot}{true}\setpropertyifundefined{compilation}{appendix}{true}\setpropertyifundefined{compilation}{affidavit}{true}\setpropertyifundefined{compilation}{clsdefineschapter}{true}}{}\@ifclassloaded{article}{\setpropertyifundefined{compilation}{titlepage}{true}\setpropertyifundefined{compilation}{abstract}{true}\setpropertyifundefined{compilation}{toc}{true}\setpropertyifundefined{compilation}{bibliography}{true}}{}\@ifclassloaded{scrbook}{\setpropertyifundefined{compilation}{frontmatter}{true}\setpropertyifundefined{compilation}{backmatter}{true}\setpropertyifundefined{compilation}{titlepage}{true}\setpropertyifundefined{compilation}{abstract}{true}\setpropertyifundefined{compilation}{toc}{true}\setpropertyifundefined{compilation}{bibliography}{true}\setpropertyifundefined{compilation}{lof}{true}\setpropertyifundefined{compilation}{lot}{true}\setpropertyifundefined{compilation}{appendix}{true}\setpropertyifundefined{compilation}{affidavit}{true}\setpropertyifundefined{compilation}{clsdefineschapter}{true}}{}\@ifclassloaded{scrreprt}{\setpropertyifundefined{compilation}{titlepage}{true}\setpropertyifundefined{compilation}{abstract}{true}\setpropertyifundefined{compilation}{toc}{true}\setpropertyifundefined{compilation}{bibliography}{true}\setpropertyifundefined{compilation}{lof}{true}\setpropertyifundefined{compilation}{lot}{true}\setpropertyifundefined{compilation}{appendix}{true}\setpropertyifundefined{compilation}{affidavit}{true}\setpropertyifundefined{compilation}{clsdefineschapter}{true}}{}\@ifclassloaded{scrartcl}{\setpropertyifundefined{compilation}{titlepage}{true}\setpropertyifundefined{compilation}{abstract}{true}\setpropertyifundefined{compilation}{toc}{true}\setpropertyifundefined{compilation}{bibliography}{true}}{}\@ifclassloaded{moderncv}{\setpropertyifundefined{compilation}{titlepage}{false}\setpropertyifundefined{compilation}{abstract}{false}\setpropertyifundefined{compilation}{toc}{false}\setpropertyifundefined{compilation}{bibliography}{false}\setpropertyifundefined{compilation}{glossaries}{false}}{}\setpropertyifundefined{compilation}{frontmatter}{false}\setpropertyifundefined{compilation}{backmatter}{false}\setpropertyifundefined{compilation}{titlepage}{false}\setpropertyifundefined{compilation}{abstract}{false}\setpropertyifundefined{compilation}{toc}{false}\setpropertyifundefined{compilation}{los}{false}\setpropertyifundefined{compilation}{bibliography}{false}\setpropertyifundefined{compilation}{lof}{false}\setpropertyifundefined{compilation}{lot}{false}\setpropertyifundefined{compilation}{lol}{false}\setpropertyifundefined{compilation}{appendix}{false}\setpropertyifundefined{compilation}{listofpublications}{false}\setpropertyifundefined{compilation}{acknowledgement}{false}\setpropertyifundefined{compilation}{affidavit}{false}\setpropertyifundefined{compilation}{clsdefineschapter}{false}\setpropertyifundefined{compilation}{acronyms}{false}\setpropertyifundefined{compilation}{glossaries}{true}\makeatother
\RequirePackageWithOption[
			fleqn 
		]
		{amsmath}

\usepackage{amsthm}
\usepackage{calc}
\ifluatex
\else
    \usepackage
		[
		]
		{amsfonts}
    \usepackage
		[
		]
		{amssymb}
\fi
\ifluatex
	\usepackage
		[
		]
		{fontspec}

\setproperty{compilation}{fontspec}{true}
	\usepackage
		[
			math-style=ISO,
			bold-style=ISO,
			nabla=upright,
		]
		{unicode-math}
	\AtEndPreamble{
        \ifthenelseproperty{compilation}{minionfonts}{
			\setmathfont{Cambria Math}
			\ifthenelseproperty{compilation}{complementglyphs}{%
				\setmathfont[range=\mathup/{num,latin,Latin,greek,Greek}]{Minion Pro}
				\setmathfont[range=\mathbfup/{num,latin,Latin,greek,Greek}]{MinionPro-Bold}
				\setmathfont[range=\mathit/{num,latin,Latin,greek,Greek}]{MinionPro-It}
				\setmathfont[range=\mathbfit/{num,latin,Latin,greek,Greek}]{MinionPro-BoldIt}
				\setmathfont[range={"0025,"002B,"002C,"002D,"002E,"2212,"2248,"007E,`±,`×,`·,`÷,`¬,`∂,`∆,`∕,`∞,`⌐,`<,`>,`=,`≥,`≤,`−}]{Minion Pro}
				\setmathfont[range={"2211,"221A}]{Latin Modern Math}
			}{}
			\renewcommand{\mathcal}[1]{{\textit{\addfontfeatures{Contextuals=Swash}#1}}}
			\newfontface{\MinionProSwash}{MinionPro-It}
					[
						Contextuals=Swash,
						Ligatures={TeX,Common,Rare,Historic,Contextual,Required}
					]
			\typeout{----- USING MINION FONTS -----}
		}{
		    \ifthenelseproperty{compilation}{libertinusfonts}{
		    	\setmainfont{Libertinus Serif}
		[
			BoldFont          = {* Bold},
  			ItalicFont        = {* Italic},
  			BoldItalicFont    = {* Bold Italic},
			SmallCapsFeatures = {Renderer=Basic},
			Ligatures         = {TeX,Common},
			Scale             = MatchLowercase,
			RawFeature=-lnum;+pnum;+onum;
		]
		    	\setsansfont{Libertinus Sans}
		[
			BoldFont          = {* Bold},
  			ItalicFont        = {* Italic},
  			BoldItalicFont    = {* -Bold Italic},
			SmallCapsFeatures = {Renderer=Basic},
			Ligatures         = {TeX,Common},
			Scale             = MatchLowercase,
			Numbers           = {Lining,Monospaced},
		]
		    	\setmonofont{Libertinus Mono}
		[
			BoldFont          = {* Bold},
  			ItalicFont        = {* Italic},
  			BoldItalicFont    = {* Bold Italic},
			SmallCapsFeatures = {Renderer=Basic},
			Ligatures         = {Common},
			Scale             = MatchLowercase,
			Numbers           = {Lining,Monospaced},
		]
		    	\setmathfont{Cambria Math}
				\ifthenelseproperty{compilation}{complementglyphs}{%
					\setmathfont[range=\mathup/{num,latin,Latin,greek,Greek}]{LibertinusSerif}
					\setmathfont[range=\mathbfup/{num,latin,Latin,greek,Greek}]{LibertinusSerif-Bold}
					\setmathfont[range=\mathit/{num,latin,Latin,greek,Greek}]{LibertinusSerif-Italic}
					\setmathfont[range=\mathbfit/{num,latin,Latin,greek,Greek}]{LibertinusSerif-BoldItalic}
					\setmathfont[range={"0025,"002B,"002C,"002D,"002E,"2212,"2248,"007E,"2208,"221D,"2A85,`±,`×,`·,`÷,`¬,`∂,`∆,`∕,`∞,`⌐,`<,`>,`=,`≥,`≤,`−,`;,}]{Libertinus Math}
					\setmathfont[range={"2211,"221A}]{Latin Modern Math}
					\setmathfont[range=\mathcal/{num,latin,Latin,greek,Greek},Contextuals=Swash]{Cambria Math}
				}{}
		    	\newfontface{\LibertinusSwash}{LibertinusSerif-Italic}
		    			[
		    				Contextuals=Swash,
		    				Ligatures={TeX,Common,Rare,Historic,Contextual,Required}
		    			]
		    	\newfontface{\LibertinusInitials}{LibertinusSerif-Initials}
		    	\typeout{----- USING LIBERTINUS FONTS -----}
		    }{
                \setmathfont{latinmodern-math.otf}
				\ifthenelseproperty{compilation}{complementglyphs}{%
                    \setmathfont[range=\setminus]{Asana Math}
				}{}
            }
        }
	}
\else
	\usepackage
		[
			T1
		]
		{fontenc}
	\usepackage
		[
			utf8
		]
		{inputenc}
	\usepackage{lmodern}
\fi

\usepackage{fontawesome5} 
\ifluatex
    \usepackage[a-3b,pdf17]{pdfx}
    \usepackage{xmpincl}

    \pdfvariable omitcidset=1
	\newcommand{\replaceNBSP}[1]{\luadirect{s,_ = string.gsub("\luatexluaescapestring{#1}", "\luatexluaescapestring{~}", " "); tex.print(-2,s)}}

    \AtBeginDocument{
    	\newwrite\metadatafile
    	\immediate\openout\metadatafile=\jobname.xmpdata
    	\immediate\write\metadatafile{\unexpanded{\Title}{\expanded{\replaceNBSP{\getproperty{document}{title}}}}}
    	\immediate\write\metadatafile{\unexpanded{\Author}{\expanded{\getproperty{author}{firstname} \getproperty{author}{familyname}}}}
    	\immediate\write\metadatafile{\unexpanded{\Subject}{\expanded{\getproperty{document}{subject}}}}
    	\immediate\write\metadatafile{\unexpanded{\Keywords}{\getproperty{document}{keywords}}}
    	\immediate\write\metadatafile{\unexpanded{\PublicationType}{\expanded{\getproperty{document}{type}}}}
    	\immediate\write\metadatafile{\unexpanded{\Doi}{\expanded{\getproperty{document}{doi}}}}
    	\immediate\closeout\metadatafile
    }
\fi

\usepackage
		[
			locale=UK,
			separate-uncertainty,
			binary-units, 
		]
		{siunitx}

\DeclareSIUnit{\arbitraryunit}{a.\,u.}

\DeclareSIUnit{\permille}{‰}

\DeclareSIUnit{\sample}{S}

\DeclareSIPrefix{\Femto}{f\kern0.1ex}{-15}

\DeclareSIUnit{\pb}{\pico\barn}
\DeclareSIUnit{\fb}{\femto\barn}
\DeclareSIUnit{\nb}{\nano\barn}

\usepackage{chemfig}

\AtEndPreamble{\usepackage{scrhack}}

\AtEndPreamble{
	\ifthenelseproperty{compilation}{pdfa}{
	}{
		\usepackage{intopdf}
		
	}
}

\usepackage{seqsplit}

\usepackage{booktabs}

\ifthenelseproperty{compilation}{fontspec}{%
	\AtBeginEnvironment{tabular}{\addfontfeatures{Numbers={Monospaced}}}
    }{}

\usepackage{xcolor} 

\PassOptionsToPackage{dvipsnames,svgnames,x11names,rgb}{xcolor}
\definecolor{darkgreen}{RGB}{0,80,0} 
\definecolor{darkred}{RGB}{80,0,0}

\usepackage{graphicx}

\usepackage{pgf}  

\usepackage
		[
			format=plain, 
			indention=1em, 
			textfont={color={black},small}, 
			width=0.925\linewidth, 
			hypcap=true, 
		]
		{caption}

\AtBeginDocument
			{
				\captionsetup
						{
                            labelfont={color=\getproperty{color}{main},small,sf,bf}, 
						}
			}
\usepackage{wrapfig}

\usepackage
		[
			hypcap=true, 
			textfont={color={black},small}, 
			width=0.925\textwidth, 
			indention=1em, 
		]
		{subcaption}

\usepackage{listings}

\definecolor{javagray}{rgb}{0.55, 0.52, 0.54} 
\definecolor{javared}{rgb}{0.6,0,0} 
\definecolor{javagreen}{rgb}{0.25,0.5,0.35} 
\definecolor{javapurple}{rgb}{0.5,0,0.35} 
\definecolor{javadocblue}{rgb}{0.25,0.35,0.75} 
\definecolor{javaLila}{RGB}{127,0,85}

\let\origthelstnumber\thelstnumber

\makeatletter
\newcommand*\Suppressnumber{%
  \lst@AddToHook{OnNewLine}{%
    \let\thelstnumber\relax%
  }%
}
\newcommand\Reactivatenumber[1]{%
  \global\c@lstnumber#1%
  \global\advance\c@lstnumber\m@ne\relax%
  \lst@AddToHook{OnNewLine}{%
  \let\thelstnumber\origthelstnumber%
  }%
}
\makeatother

\makeatletter
\lst@InputCatcodes
\def\lst@DefEC{%
 \lst@CCECUse \lst@ProcessLetter
^^80^^81^^82^^83^^84^^85^^86^^87^^88^^89^^8a^^8b^^8c^^8d^^8e^^8f%
  ^^90^^91^^92^^93^^94^^95^^96^^97^^98^^99^^9a^^9b^^9c^^9d^^9e^^9f%
  ^^a0^^a1^^a2^^a3^^a4^^a5^^a6^^a7^^a8^^a9^^aa^^ab^^ac^^ad^^ae^^af%
  ^^b0^^b1^^b2^^b3^^b4^^b5^^b6^^b7^^b8^^b9^^ba^^bb^^bc^^bd^^be^^bf%
  ^^c0^^c1^^c2^^c3^^c4^^c5^^c6^^c7^^c8^^c9^^ca^^cb^^cc^^cd^^ce^^cf%
  ^^d0^^d1^^d2^^d3^^d4^^d5^^d6^^d7^^d8^^d9^^da^^db^^dc^^dd^^de^^df%
  ^^e0^^e1^^e2^^e3^^e4^^e5^^e6^^e7^^e8^^e9^^ea^^eb^^ec^^ed^^ee^^ef%
  ^^f0^^f1^^f2^^f3^^f4^^f5^^f6^^f7^^f8^^f9^^fa^^fb^^fc^^fd^^fe^^ff%
	^^^^03b8^^^^03c8^^^^03b7^^^^03bc^^^^03c3^^^^03b1^^^^03a9^^^^03b6%
	^^^^03c9^^^^03b4^^^^03c0%
^^00}
\lst@RestoreCatcodes
\makeatother

\RequirePackageWithOption{hyperref}

\hypersetup
		{
			pdfencoding=unicode,
			pdfpagemode=UseOutlines,
			bookmarksopenlevel=0,
			pdfpagelayout=TwoPageRight,
			hidelinks,
			pdfduplex=DuplexFlipLongEdge,
		}

\ifluatex
	\RequirePackageWithOption
			[
				activate={true,nocompatibility}, 
				final, 
				tracking=true,
				factor=1100, 
				stretch=10, 
				shrink=10, 
			]
			{microtype}
\else
	\RequirePackageWithOption
			[
				activate={true,nocompatibility}, 
				final, 
				tracking=true,
				kerning=true, 
				spacing=true, 
				factor=1100, 
				stretch=10, 
				shrink=10, 
			]
			{microtype}
\fi

\setproperty{compilation}{microtype}{true}

\SetTracking[no ligatures]{encoding={*}, shape=sc}{40} 
\usepackage{epigraph}

\usepackage
		[
			math,
			toc,
		]
		{blindtext}

\usepackage
		[
			noabbrev
		]
		{cleveref}

\crefname{line}{line}{lines}
\creflabelformat{line}{#2#1#3}

\crefname{reaction}{reaction}{reactions}
\creflabelformat{reaction}{#2(#1)#3}

\crefname{listing}{code}{codes}

\usepackage
		[
			strict=true, 
			autostyle=true, 
			german=guillemets, 
		]
		{csquotes}

\setquotestyle{german}

\usepackage{lettrine} 

\setpropertyifundefined{biblatex}{style}{numeric-comp}\setpropertyifundefined{biblatex}{use-url}{false}\setpropertyifundefined{biblatex}{use-doi}{false}\usepackage
[
hyperref=true,
backend=biber,
url=\getproperty{biblatex}{use-url},
abbreviate=false,
isbn=false,
doi=\getproperty{biblatex}{use-doi},
backref=false,
urldate=short,
labeldate=year,
date=year,
eprint=false,
maxbibnames=3,
giveninits=true,
block=none,
sorting=none,
defernumbers=true,
style=numeric-comp,
]
{biblatex}\AtEveryBibitem{\clearfield{pages}}
\AtEveryBibitem{\clearfield{volume}}
\AtEveryBibitem{\clearfield{number}}
\renewbibmacro{in:}{}\makeatletter
\@ifpackageloaded{seqsplit}{\DeclareFieldFormat{eid}{Art.\addnbspace\seqsplit{#1}}
}{}\makeatother
\DeclareSourcemap{\maps[datatype=bibtex]{ \map{ \step[fieldsource=doi,final] \step[fieldset=url,null] }}
}
\DeclareFieldFormat{journaltitle}{\mkbibemph{#1},}
\DeclareFieldFormat[inbook,thesis]{title}{\mkbibemph{#1}\addperiod}
\DeclareFieldFormat[article,misc]{title}{\enquote{#1}}
\DeclareFieldFormat[article,inproceedings,incollection]{volume}{Vol.~#1}\DeclareFieldFormat{edition}{\ifinteger{#1}{\mkbibordedition{#1}\addnbthinspace{}ed.}{#1\isdot}}
\DeclareFieldFormat{eprint:urn}{\textsc{URN}\addcolon\space\ifhyperref{\href{http://www.nbn-resolving.org/#1}{\nolinkurl{#1}}}{\nolinkurl{#1}}}

\DeclareSortingTemplate{ndymdt}{\sort{ \field{presort}}
\sort[final]{\field{sortkey}}
\sort[direction=descending]{\field{sortyear}\field{year}\literal{9999}}
\sort[direction=descending]{\field[padside=left,padwidth=2,padchar=0]{month}\literal{99}}
\sort[direction=descending]{\field[padside=left,padwidth=2,padchar=0]{day}\literal{99}}
}
\ifluatex
\def\getAuthorBibCv{\FirstChar{\getproperty{author}{firstname}}.~\getproperty{author}{familyname}}
\else
\def\getAuthorBibCv{\getproperty{author}{firstname}~\getproperty{author}{familyname}}
\fi
\newboolean{isCoauthorList}\setboolean{isCoauthorList}{false}\DefineBibliographyStrings{english}{andothers={ \ifthenelse{ \boolean{isCoauthorList}}{et\addabbrvspace al\adddot\addcomma\addspace amongst them \textbf{\textsc{\getAuthorBibCv}}
}{ et\addabbrvspace al\adddot}}
}
\DeclareBibliographyCategory{dontbib}\let\printbibliographyold\printbibliography
\renewcommand{\printbibliography}[1][]{\typeout{----- printbibliography ------}\printbibliographyold[notcategory=dontbib,#1]
}
\makeatletter
\newcommand{\tempmaxup}[1]{\def\blx@maxcitenames{99}#1}
\DeclareCiteCommand{\fullcitecontribution}[\tempmaxup]{\usebibmacro{prenote}\addtocategory{dontbib}{\thefield{entrykey}}
}{\textbf{\usebibmacro{maintitle+title}}
\newline\nopunct\newblock
\usebibmacro{author}\newline\nopunct\newblock
\usebibmacro{journal+issuetitle},\usebibmacro{doi+eprint+url} \usebibmacro{addendum+pubstate}}{\multicitedelim }{\usebibmacro{postnote}}
\makeatother

\ifKOMA
    \usepackage
    		[
    			automark,
    			headsepline, 
    		]
    		{scrlayer-scrpage}
    
    \AtBeginDocument{%
		\addtokomafont{pagehead}{\color{\getproperty{color}{main}}}
		\ifthenelseproperty{compilation}{clsdefineschapter}{%
			\addtokomafont{chapter}{\color{\getproperty{color}{main}}}
		}{}
		\addtokomafont{section}{\color{\getproperty{color}{main}}}
		\addtokomafont{subsection}{\color{\getproperty{color}{main}}}
		\addtokomafont{subsubsection}{\color{\getproperty{color}{main}}}
		\addtokomafont{paragraph}{\color{\getproperty{color}{main}}}

		\addtokomafont{footnoterule}{\color{\getproperty{color}{secondary}}}
		\addtokomafont{headsepline}{\color{\getproperty{color}{secondary}}}

    }
\fi

\usepackage{bookmark}

\bookmarksetup
			{
				open,
				addtohook={
					\ifnum\bookmarkget{level}<1
						\bookmarksetup{bold}
					\fi
				},
			}

\PassOptionsToPackage{automake=immediate}{glossaries-extra}

\usepackage%
		[
			nonumberlist, 
			acronym, 
			toc, 
		]
		{glossaries-extra}

\usepackage{glossary-mcols}

\setabbreviationstyle[acronym]{long-short}

\newglossary[slg]{symbols}{syi}{syg}{List of Symbols} 

\makeglossaries

\newglossarystyle{customGlossaryStyleForListOfSymbols}{%
    \renewenvironment{theglossary}{%
        \begin{longtable}{p{0.12\textwidth}p{\glsdescwidth}p{\glspagelistwidth}}
    }{%
        \end{longtable}
    }
    
    \renewcommand*{\glsgroupheading}[1]{}

}

\newglossarystyle{customListOfSymbols}{
    \setglossarystyle{customGlossaryStyleForListOfSymbols}
    
    \renewenvironment{theglossary}{%
        \begin{longtable}{p{0.12\textwidth}p{\glsdescwidth}p{\glspagelistwidth}}
    }{%
        \end{longtable}
    }
}

\newglossarystyle{customListOfAbbreviations}{%
    \renewenvironment{theglossary}{%
        \begin{longtable}{lp{\glsdescwidth}}
    }{%
        \end{longtable}
    }
    
    \renewcommand*{\glsgroupheading}[1]{}

}

\makeatletter
\newglossarystyle{customIndex}{
    \renewenvironment{theglossary}{%
        \setlength{\parindent}{0pt}
        \setlength{\parskip}{0pt plus 0.3pt}
        \let\item\@idxitem
    }{%
    }
    
    \renewcommand*{\glsgroupheading}[1]{}
    \renewcommand*{\glossaryentryfield}[5]{%
    \item\glstarget{##1}{##2}
        \ifx\relax##4\relax
        \else
            \space(##4)
        \fi
        \dotfill ##3\glspostdescription \space ##5
    }
    \renewcommand*{\glossarysubentryfield}[6]{%
        \ifcase##1\relax
        \item
        \or
            \subitem
        \else
            \subsubitem
        \fi
        \glstarget{##2}{##3}
        \ifx\relax##5\relax
        \else
            \space(##5)
        \fi
        \dotfill ##4\glspostdescription\space ##6
    }
    
    \renewcommand*{\glsgroupheading}[1]{%
        \item\textbf{\glsgetgrouptitle{##1}}\indexspace
    }
}
\makeatother

\newglossarystyle{documentationLabelStyle}{
    \renewenvironment{theglossary}{%
        \begin{longtable}{p{.2\textwidth}p{.3\textwidth}p{.5\textwidth}}
    }{%
        \end{longtable}
    }%

}

\newglossarystyle{documentationSymbolsStyle}{
    \renewenvironment{theglossary}{%
        \begin{longtable}{p{0.12\textwidth}p{.2\textwidth}p{.48\textwidth}p{.2\textwidth}}
    }{%
        \end{longtable}
    }

}

\newglossarystyle{publicationLabelStyle}{
	\renewenvironment{theglossary}{%
		\begin{longtable}{p{.2\textwidth}p{.8\textwidth}}
		}{%
		\end{longtable}
	}%

}

\newglossarystyle{publicationSymbolsStyle}{
}

\usepackage{scrtime}

\usepackage{ragged2e}

\usepackage{marginnote}

\DeclareDocumentCommand{\myMarginnote} 
					{
				 		O{0cm} O{c} m 
				 	}{
						\marginnote
								{
									\ifthispageodd
											{
												\RaggedRight 
											}{
												\RaggedLeft 
											}
									\raisebox{#1}[#1]
											{
												\begin{minipage}[#2]{\marginparwidth}
													\RaggedRight 
													\color{\getMainColor}
													\lineskiplimit=-\maxdimen
													\normalfont\sffamily
													#3\end{minipage}}
											}
					}

\ifluatex
	\usepackage[luatex]{pdflscape}
\else
\fi

\usepackage
		[
			final, 
		]
		{pdfpages}

\pretocmd{\includepdf}{%
    \ifthenelseproperty{compilation}{externalize}{%
        \tikzset{external/optimize=false}%
    }{}
}{}{}

\makeatletter
\@ifpackageloaded{cleveref}{%
    \newcounter{articlenumber}
    \crefname{articlenumber}{article}{articles}
    \Crefname{articlenumber}{Article}{Articles}

}
\makeatother

\usepackage[
    inline, 
]{enumitem}

\setlist[description]{leftmargin=*,style=sameline}

\ifluatex

\usepackage{luacode}

\usepackage
		[
			draft, 
		]
		{impnattypo}

\usepackage{selnolig}

\usepackage{pdftexcmds}
\makeatletter
\let\pdfstrcmp\pdf@strcmp
\makeatother
\else
\fi
\input{preamble_defs.tex}\setproperty{document}{subject}{Paper}\setproperty{document}{titlehead}{}\setproperty{document}{title}{Improving ideal MHD equilibrium accuracy with physics-informed neural networks}\setproperty{author}{firstname}{Timo}\setproperty{author}{familyname}{Thun}\setproperty{author}{affiliationindices}{1}\setproperty{coauthor1}{firstname}{Andrea}\setproperty{coauthor1}{familyname}{Merlo}\setproperty{coauthor1}{affiliationindices}{2}\setproperty{coauthor2}{firstname}{Rory}\setproperty{coauthor2}{familyname}{Conlin}\setproperty{coauthor2}{affiliationindices}{3}\setproperty{coauthor3}{firstname}{Dario}\setproperty{coauthor3}{familyname}{Panici}\setproperty{coauthor3}{affiliationindices}{4}\setproperty{coauthor4}{firstname}{Daniel}\setproperty{coauthor4}{familyname}{Böckenhoff}\setproperty{coauthor4}{affiliationindices}{1}\setproperty{affiliations}{1}{Max-Planck-Institute for Plasma Physics,17491 Greifswald,Germany}\setproperty{affiliations}{2}{Proxima Fusion GmbH,81369 Munich,Germany}\setproperty{affiliations}{3}{Institute for Research in Electronics \& Applied Physics,University of Maryland,MD 20742,USA}\setproperty{affiliations}{4}{Mechanical and Aerospace Engineering,Princeton University,NJ 08544,USA}\usepackage[final]{changes}\colorlet{Changes@Color}{red}\usepackage{placeins}\usepackage{circuitikz}\setcommentmarkup{\footnote{#1}}
\definechangesauthor[name={Timo Thun},color=red]{TT}

\newcommand{\vect}[1]{\ensuremath{\vec{#1}}}
\newcommand{\matr}[1]{\ensuremath{\makebf{#1}}}
\makeatletter
\newcommand{\Grad}[2][\@nil]{\def\tmp{#1}\ifx\tmp\@nnil
\ensuremath{\vect{\nabla} #2}
\else
\ensuremath{\vect{\nabla}_{\!\!#1} #2}
\fi}
\makeatother

\newcommand{\dx}[3][\empty]
{\if{#1}\equal{\empty}\frac{\mathrm{d}#2}{\mathrm{d}#3}
\else
\frac{\mathrm{d}^{#1}#2}{\mathrm{d}#3^{#1}}
}
\newcommand{\pdx}[3][\empty]
{\if{#1}\equal{\empty}\frac{\partial#3}{\partial#2}\else
\frac{\partial^{#1}#3}{\partial#2^{#1}}
}
\makeatletter
\providecommand*{\diff}{\@ifnextchar^{\DIfF}{\DIfF^{}}}
\def\DIfF^#1{\mathop{\mathrm{\mathstrut d}}\nolimits^{#1}\gobblespace}
\def\gobblespace{\futurelet\diffarg\opspace}\def\opspace
{\let\DiffSpace\! \ifx\diffarg( \let\DiffSpace\relax \else \ifx\diffarg[ \let\DiffSpace\relax \else \ifx\diffarg\{ \let\DiffSpace\relax \fi \fi \fi \DiffSpace }\makeatother

\DeclareMathOperator*{\argmin}{arg\,min}

\newcommand{\tightoverset}[2]{\mathop{#2}\limits^{\vbox to -.5ex{\kern-0.75ex\hbox{$\! #1$}\vss}}}

\DeclareSIUnit[number-unit-product=\,]{\permille}{\textperthousand}
\newacronym{statistics}{Statistics}{\nopostdesc}\newacronym[parent=statistics,plural=MCs]{MC}{MC}{Monte Carlo}\newacronym[parent=statistics,plural=MCMCs]{MCMC}{MCMC}{Markov chain Monte Carlo}\newacronym[parent=statistics]{MAP}{MAP}{maximum a posteriori}\newacronym[parent=statistics,plural=LAs,longplural={Laplace approximations}]{LA}{LA}{Laplace approximation}\newacronym[parent=statistics,first=kernel density estimate (KDE) \cite{Parzen1962,Silverman1986},plural=KDEs,firstplural=kernel density estimates (KDEs) \cite{Parzen1962,Silverman1986},longplural={kernel density estimates}]{KDE}{KDE}{kernel density estimate}\newacronym[parent=statistics,plural=PDFs,longplural={probability density functions}]{PDF}{PDF}{probability density function}\newacronym[parent=statistics,plural=GPs,longplural={Gaussian processes}]{GP}{GP}{Gaussian process}\newacronym[parent=statistics,plural=LGIs,longplural={linear Gaussian inversions}]{LGI}{LGI}{linear Gaussian inversion}\newacronym[parent=statistics,plural=LSs,longplural={least-squares}]{LS}{LS}{least-square}\newacronym[parent=statistics]{MaLi}{Max-$\mathcal{L}$}{maximum likelihood}\newacronym[parent=statistics,plural=iid,longplural={independent and identically distributed}]{iid}{iid}{independent and identically distributed}\newacronym[parent=statistics]{FWHM}{FWHM}{full-width half-maximum}\newacronym[parent=statistics]{HDI}{HDI}{highest density interval}\newacronym[parent=statistics]{MHA}{MHA}{Metropolis-Hastings algorithm}\newacronym[parent=statistics]{KS}{KS}{Kolmogorov–Smirnov}\newglossaryentry{PearsonCorrelationSample}{parent=statistics,name=\ensuremath{r_{\text{xy}}},
type=symbols,
sort=correlation,
description={Pearson sample correlation coefficient},
symbol={},
}
\newacronym[parent=statistics,plural=HPs,longplural={hyper-parameters}]{HP}{HP}{hyper-parameter}\newacronym[parent=statistics,plural=PCA]{PCA}{PCA}{principal component analysis}\newacronym{dynamicalSystems}{Dynamical systems}{\nopostdesc}\newacronym[parent=dynamicalSystems]{KAM}{KAM}{Kolmogorov–Arnold–Moser}\newacronym{methods}{Methods}{\nopostdesc}\newacronym[parent=methods]{FP}{FP}{Function parametrization}\newacronym[parent=methods]{MISO}{MISO}{multiple-input single-output}\newacronym[parent=methods]{MIMO}{MIMO}{multiple-input multiple-output}\newacronym{functionalAnalysis}{Functional analysis}{\nopostdesc}\newacronym[parent=functionalAnalysis,plural=FCs,longplural={Fourier coefficients}]{FC}{FC}{Fourier coefficient}\newacronym{differentialEquations}{Differential Equations}{\nopostdesc}\newacronym[parent=differentialEquations,plural=ODEs,longplural={Ordinary Differential Equations}]{ODE}{ODE}{Ordinary Differential Equation}\newacronym[parent=differentialEquations,plural=PDEs,longplural={Partial Differential Equations}]{PDE}{PDE}{Partial Differential Equation}\newacronym{equations}{Equations}{\nopostdesc}\newacronym[parent=equations]{LHS}{LHS}{left-hand side}\newacronym[parent=equations]{RHS}{LHS}{right-hand side}
\newacronym{GeneralPhysics}{General Physics}{\nopostdesc}\newacronym[parent=GeneralPhysics]{IR}{IR}{Infrared}\newacronym[parent=GeneralPhysics]{EM}{EM}{electro-magnetic}\newacronym{Organisational}{Organisational}{\nopostdesc}\newacronym[parent=Organisational]{OP}{OP}{operation phase}
\newacronym[parent=Organisational]{KKS}{KKS}{\emph{Kraftwerk-Kennzeichensystem} Reference Designation System for Power Plants}
\newacronym[parent=Organisational]{PLM}{PLM}{Project Lifecycle Management}\newacronym{plasmaTheory}{Plasma Theory}{\nopostdesc}\newacronym[parent=plasmaTheory,plural=tokamaks]{tokamak}{tokamak}{тороидальная камера в магнитных катушках}\newacronym[parent=plasmaTheory]{MHD}{MHD}{magnetohydrodynamic}\newacronym[parent=plasmaTheory]{LCFS}{LCFS}{last closed flux surface}\newacronym[parent=plasmaTheory]{LCMS}{LCMS}{last closed magnetic surface}\newacronym[parent=plasmaTheory]{ISS95}{ISS95}{international stellarator scaling 1995}\newacronym[parent=plasmaTheory]{ISS04}{ISS04}{international stellarator scaling 2004}\newacronym[parent=plasmaTheory]{NTM}{NTM}{neoclassical tearing modes}\newacronym[parent=plasmaTheory]{HPP}{HPP}{heat pulse propagation}\newacronym{plasmaPhysics}{Plasma Physics}{\nopostdesc}\newacronym[parent=plasmaPhysics]{Omode}{O~mode}{ordinary mode}\newacronym[parent=plasmaPhysics]{Xmode}{X~mode}{extraordinary mode}\newacronym{X2mode}{X2~mode}{second harmonic extraordinary mode}\newacronym{O2mode}{O2~mode}{second harmonic ordinary mode}\newacronym{X3mode}{X3~mode}{third harmonic extraordinary mode}\newacronym[parent=plasmaPhysics]{EBW}{EBW}{electron Bernstein wave}\newacronym[parent=plasmaPhysics]{TM}{TM}{transverse magnetic}\newacronym[parent=plasmaPhysics]{TEM}{TEM}{transverse electromagnetic}\newacronym[parent=plasmaPhysics,plural=EEDFs,firstplural=electron energy distribution functions (EEDFs)]{EEDF}{EEDF}{electron energy distribution function}\newacronym[parent=plasmaPhysics]{ECE}{ECE}{electron cyclotron emission}\newacronym[parent=plasmaPhysics]{LFS}{LFS}{low field side}\newacronym[parent=plasmaPhysics]{HFS}{HFS}{high field side}\newacronym[parent=plasmaPhysics]{ELM}{ELM}{edge localized mode}\newacronym[parent=plasmaPhysics]{MARFE}{MARFE}{multifaceted asymmetric radiation from the edge}\newacronym[parent=plasmaPhysics]{Serpens}{Serpens}{self-regulated plasma edge beneath the last-closed-flux-surface}\newacronym[parent=plasmaPhysics,plural=SOLs,firstplural=scrape-off layers]{SOL}{SOL}{scrape-off layer}\newacronym[parent=plasmaPhysics]{ECRH}{ECRH}{electron cyclotron resonance heating}\newacronym[parent=plasmaPhysics]{ECCD}{ECCD}{electron cyclotron current drive}\newacronym[parent=plasmaPhysics]{NBCD}{NBCD}{neutral beam current drive}\newacronym[parent=plasmaPhysics]{ICRH}{ICRH}{ion cyclotron radiation heating}\newacronym[parent=plasmaPhysics]{FL}{FL}{field line}\newacronym[parent=plasmaPhysics]{PFR}{PFR}{private flux region}\newacronym[parent=plasmaPhysics,firstplural=reversed-field pinches]{RFP}{RFP}{reversed-field pinch}\newacronym[parent=plasmaPhysics]{FRC}{FRC}{field-reversed configuration}\newacronym[parent=plasmaPhysics]{see}{\textsc{SEE}}{secondary electron emission}\newacronym[parent=plasmaPhysics]{SE}{\textsc{SE}}{sheath edge}\newacronym[parent=plasmaPhysics]{SL}{SL}{strike line}\newacronym[parent=plasmaPhysics]{PWI}{PWI}{plasma wall interaction}\newacronym{fusionReactorComponent}{Nuclear Fusion Machine Components}{\nopostdesc}\newacronym[parent=fusionReactorComponent,plural=PFCs,firstplural=plasma facing components (PFCs)]{PFC}{PFC}{plasma facing component}\newacronym[parent=fusionReactorComponent,]{VV}{VV}{vaccum vessel}\newacronym[parent=fusionReactorComponent,plural=IVCs,firstplural=in-vessel components (IVCs)]{IVC}{IVC}{in-vessel component}\newacronym[parent=fusionReactorComponent,plural=TDUs,firstplural=test divertor units (TDUs)]{TDU}{TDU}{test divertor unit}\newacronym[parent=fusionReactorComponent,plural=HHFDs,firstplural=high heatflux divertors (HHFDs)]{HHFD}{HHFD}{high heatflux divertor}\newacronym[parent=fusionReactorComponent,]{TDUSE}{TDUSE}{test divertor unit scraper element}\newacronym[parent=fusionReactorComponent,]{PG}{PG}{Pumping Gap}\newacronym[parent=fusionReactorComponent,]{TE}{TE}{Target Element}\newacronym[parent=fusionReactorComponent]{NBI}{NBI}{neutral beam injection}\newacronym[parent=fusionReactorComponent,plural=PIs,firstplural=pellet injections (PIs)]{PI}{PI}{pellet injection}\newacronym{plasmaTheoryCode}{Plasma Theory Codes}{\nopostdesc}\newacronym[parent=plasmaTheoryCode]{FLD}{FLD}{field line diffusion}\newacronym[parent=plasmaTheoryCode]{FLT}{FLT}{field line tracing}\newacronym[parent=plasmaTheoryCode]{LOWENDEL}{LOWENDEL}{``loads on Wendelstein''}\newacronym[parent=plasmaTheoryCode]{VMEC}{VMEC}{variational moments equilibrium code}\newacronym[parent=plasmaTheoryCode]{EXTENDER}{EXTENDER}{virtual casing principle code}\newacronym[parent=plasmaTheoryCode]{PIES}{PIES}{Princeton iterative equilibrium solver}\newacronym[parent=plasmaTheoryCode]{SPEC}{SPEC}{stepped-pressure equilibrium code}\newacronym[parent=plasmaTheoryCode]{EMC3E}{EMC3-Eirene}{Edge Monte Carlo 3D code coupled to the neutral transport code EIRENE}\newacronym[parent=plasmaTheoryCode]{TRAVIS}{TRAVIS}{TRAcing VISualized}\newacronym[parent=plasmaTheoryCode]{SPECE}{SPECE}{solver for plasma electron cyclotron emission}\newacronym[parent=plasmaTheoryCode]{EFIT}{EFIT}{equilibrium fitting}\newacronym[parent=plasmaTheoryCode]{Theo}{THEODOR}{Thermal Energy onto Divertor}\newacronym[parent=plasmaTheoryCode]{STRAHL}{STRAHL}{Radial impurity transport and emission code}\newacronym[parent=plasmaTheoryCode]{ROSE}{ROSE}{ROSE Optimises Stellarator Equilibria}\newacronym[parent=plasmaTheoryCode]{V3FIT}{V3FIT}{three-dimensional equilibrium reconstruction}\newacronym[parent=plasmaTheoryCode]{SIMPLE}{SIMPLE}{symplectic integration methods for particle loss estimation}\newacronym[parent=plasmaTheoryCode]{IDA}{IDA}{integrated data analysis}\newacronym[parent=plasmaTheoryCode]{BSM}{BSM}{Bayesian scientific modeling}\newacronym[parent=plasmaTheoryCode]{SBI}{SBI}{simulation-based inference}\newacronym{database}{Databases}{\nopostdesc}\newacronym[parent=database]{ADAS}{ADAS}{Atomic Data and Analysis Structure}\newacronym{plasmaDiagnostic}{Plasma Diagnostics}{\nopostdesc}\newacronym[parent=plasmaDiagnostic]{MPM}{MPM}{Multi-purpose manipulator}\newacronym[parent=plasmaDiagnostic]{TS}{TS}{Thomson scattering}\newacronym[parent=plasmaDiagnostic]{CTS}{CTS}{collective Thomson scattering}\newacronym[parent=plasmaDiagnostic]{XICS}{XICS}{X-ray imaging crystal spectroscopy}\newacronym[parent=plasmaDiagnostic]{CXRS}{CXRS}{charge exchange recombination spectroscopy}\newacronym[parent=plasmaDiagnostic]{LiBES}{LiBES}{lithium beam emission spectroscopy}\newacronym[parent=plasmaDiagnostic]{DCI}{DCI}{deuterium cyanide laser interferometry}\newacronym[parent=plasmaDiagnostic,plural=RFMs,firstplural=reflectometers (RFMs)]{RFM}{RFM}{reflectometer}\newacronym[parent=plasmaDiagnostic]{LP}{LP}{Langmuir probe}\newacronym[parent=plasmaDiagnostic,sort=Ha-Camera]{HAC}{$H_\alpha$-C}{$H_\alpha$-Camera}\newacronym[parent=plasmaDiagnostic,sort=Ha-Spectrometer]{HAS}{$H_\alpha$-S}{$H_\alpha$-Spectrometer}\newacronym[parent=plasmaDiagnostic]{IRC}{IRC}{Infrared-Camera}\newacronym[parent=plasmaDiagnostic]{HB}{HeBS}{Helium beam spectroscopy}\newacronym[parent=plasmaDiagnostic]{Hexos}{HEXOS}{High efficiency extreme ultraviolet overview spectrometer}\newacronym[parent=plasmaDiagnostic]{DI}{DI}{dispersion interferometer}\newacronym[parent=plasmaDiagnostic]{MPI}{MPI}{Martin Puplett interferometer}\newacronym[parent=plasmaDiagnostic]{FTS}{FTS}{Fourier transform spectroscopy}\newacronym{gasDiagnostic}{Gas Diagnostics}{\nopostdesc}\newacronym[parent=gasDiagnostic]{DRGA}{DRGA}{Diagnostic residual gas analyser}\newacronym[parent=gasDiagnostic]{LIF}{LIF}{Laser induced fluorescence}\newacronym{fusionApproach}{Nuclear Fusion Approaches}{\nopostdesc}\newacronym[parent=fusionApproach]{MCF}{MCF}{magnetic confinement fusion}\newacronym[parent=fusionApproach]{ICF}{ICF}{inertial confinement fusion}\newacronym[parent=fusionApproach]{MIF}{MIF}{magneto-inertial fusion}\newacronym{fusionExperiment}{Nuclear Fusion Experiments}{\nopostdesc}\newacronym[parent=fusionExperiment,sort=W7]{W7}{\protect\mbox{W7}}{Wendelstein~\protect\mbox{7}}
\newacronym[parent=fusionExperiment,sort=W7-A]{W7A}{\protect\mbox{W7-A}}{Wendelstein~\protect\mbox{7-A}}
\newacronym[parent=fusionExperiment,sort=W7-AS]{W7AS}{\protect\mbox{W7-AS}}{Wendelstein~\protect\mbox{7-AS}}
\newacronym[parent=fusionExperiment,sort=W7-X]{W7X}{\protect\mbox{W7-X}}{Wendelstein~\protect\mbox{7-X}}
\newacronym[parent=fusionExperiment]{NCSX}{NCSX}{National Compact Stellarator Experiment}\newacronym[parent=fusionExperiment]{CNT}{CNT}{Columbia Non-neutral Torus}\newacronym[parent=fusionExperiment]{CTH}{CTH}{Compact Toroidal Hybrid}\newacronym[parent=fusionExperiment,sort=LHD]{LHD}{\protect\mbox{LHD}}{large helical device}\newacronym[parent=fusionExperiment]{WEGA}{WEGA}{Wendelstein Experiment in Greifswald für die Ausbildung}\newacronym[parent=fusionExperiment]{HSX}{HSX}{Helically Symmetric Experiment}\newacronym[parent=fusionExperiment]{JET}{JET}{Joint European Torus}\newacronym[parent=fusionExperiment]{MAST}{MAST}{Mega-Ampere Spherical Tokamak}\newacronym[parent=fusionExperiment]{AUG}{AUG}{ASDEX Upgrade}\newacronym[parent=fusionExperiment]{ASDEX}{ASDEX}{axialsymmetrisches Divertor-Experiment}\newacronym[parent=fusionExperiment]{EAST}{EAST}{experimental advanced superconducting tokamak}\newacronym[parent=fusionExperiment]{ITER}{ITER}{lat.~\textit{the way} (formerly International Thermonuclear Experimental Reactor)}
\newacronym[parent=fusionExperiment]{AlcCM}{Alcator C-Mod}{Alto Campo Toro C-Mod}\newacronym[parent=fusionExperiment,sort=D]{D3D}{\mbox{DIII-D}}{Doublet III-D}\newacronym[parent=fusionExperiment,sort=JT60]{JT60}{\mbox{JT-60}}{Japan Torus-60}\newacronym[parent=fusionExperiment,sort=JT60SA]{JT60SA}{\mbox{JT-60SA}}{Japan Torus-60 Super Upgrade}\newacronym[parent=fusionExperiment,sort=KSTAR]{KSTAR}{\mbox{KSTAR}}{Korea Superconducting Tokamak Advanced Research}\newacronym[parent=fusionExperiment,sort=WEST]{WEST}{\mbox{WEST}}{Tungsten Environment in Steady-state Tokamak (formerly Tore Supra)}\newacronym[parent=fusionExperiment,sort=TCV]{TCV}{\mbox{TCV}}{Tokamak à configuration variable}\newacronym[parent=fusionExperiment,sort=TEXT]{TEXT}{\mbox{TEXT}}{Texas Tokamak}\newacronym[parent=fusionExperiment,sort=TEXTU]{TEXTU}{\mbox{TEXT-U}}{Texas Tokamak - Upgrade}\newacronym[parent=fusionExperiment,sort=Textor]{Textor}{\mbox{Textor}}{Tokamak Experiment for Technology Oriented Research}\newacronym[parent=fusionExperiment,sort=Helias]{Helias}{\mbox{Helias}}{helical advanced stellarator}\newacronym{stellaratorSymmetry}{Stellarator Symmetries}{\nopostdesc}\newacronym[parent=stellaratorSymmetry,sort=QO]{QO}{QO}{quasi-omnigeneous}\newacronym[parent=stellaratorSymmetry,sort=QI]{QI}{QI}{quasi-isodynamic}\newacronym[parent=stellaratorSymmetry,sort=QS]{QS}{QS}{quasi-symmetry}\newacronym[parent=stellaratorSymmetry,sort=QA]{QA}{QA}{quasi-axisymmetric}\newacronym[parent=stellaratorSymmetry,sort=QH]{QH}{QH}{quasi-helical}\newacronym[parent=stellaratorSymmetry,sort=QP]{QP}{QP}{quasi-poloidal}\newacronym{plasmaLab}{Plasma Physics Laboratories}{\nopostdesc}\newacronym[parent=plasmaLab]{PPPL}{PPPL}{Princeton Plasma Physics Laporatory}\newacronym[parent=plasmaLab]{IPP}{IPP}{Max-Planck Institute for Plasma Physics}

\newacronym[sort=W7-X]{magneticConfig}{\mbox{W7-X} Magnetic Configurations}{\nopostdesc}\newacronym[parent=magneticConfig]{EIM}{\textsc{EIM}}{'Standard' magnetic configuration in the centre of accessible space}\newacronym[parent=magneticConfig]{DBM}{\textsc{DBM}}{Low iota magnetic configuration}\newacronym{nuclearReaction}{Nuclear Reactions}{\nopostdesc}\newacronym[parent=nuclearReaction]{DT}{DT}{Deuterium-Tritium}
\newglossaryentry{physics}{name={General physics quantitites},
type=symbols,
description={\nopostdesc},
symbol={},
}
\newglossaryentry{n}{parent=physics,name=\ensuremath{n},
type=symbols,
sort=density,
description={Particle density,$n=N / V$},
symbol={\si{\per\cubic\meter}}
}
\newglossaryentry{T}{parent=physics,name=\ensuremath{T},
type=symbols,
sort=temperature,
description={Temperature},
symbol={\si{\eV}}
}
\newglossaryentry{energy}{parent=physics,name=\ensuremath{W},
type=symbols,
sort=energy,
description={Plasma kinetic energy},
symbol={\si{\kg\square\meter\per\square\second}}
}
\newglossaryentry{cs}{parent=physics,name=\ensuremath{\sigma},
type=symbols,
sort=cross-section,
description={Reaction cross-section},
symbol={\si{\meter\square}}
}
\newglossaryentry{v}{parent=physics,name=\ensuremath{\vect{v}},
type=symbols,
sort=velocity,
description={Particle velocity},
symbol={\si{\meter\per\second}}
}
\newglossaryentry{me}{parent=physics,name=\ensuremath{m_{\text{e}}},
type=symbols,
sort=mass,
description={Electron mass},
symbol={\si{\kilo\gram}},
}
\newglossaryentry{mp}{parent=physics,name=\ensuremath{m_{\text{p}}},
type=symbols,
sort=mass,
description={Proton mass},
symbol={\si{\kilo\gram}},
}

\newglossaryentry{B}{parent=physics,name=\ensuremath{\vect{B}},
type=symbols,
sort=electromagnetism,
description={Magnetic field},
symbol={\si{\tesla}=\si{\kg\per\ampere\per\square\second}}
}
\newglossaryentry{E}{parent=physics,name=\ensuremath{\vect{E}},
type=symbols,
sort=electromagnetism,
description={Electric field},
symbol={\si{\kg\meter\per\ampere\per\cubic\second}}
}
\newglossaryentry{chargedensity}{parent=physics,name=\ensuremath{\rho},
type=symbols,
sort=electromagnetism,
description={Electric charge density},
symbol={\si{\ampere\second\per\cubic\meter}}
}
\newglossaryentry{currentdensity}{parent=physics,name=\ensuremath{\vect{j}},
type=symbols,
sort=electromagnetism,
description={Electric current density},
symbol={\si{\ampere\per\square\meter}}
}
\newglossaryentry{resistance}{parent=physics,name=\ensuremath{R},
type=symbols,
sort=impedance,
description={Electric resistance},
symbol={\si{\ohm}}
}
\newglossaryentry{inductance}{parent=physics,name=\ensuremath{L},
type=symbols,
sort=inductance,
description={Inductance},
symbol={\si{\henry}=\si{\kg\square\meter\per\square\ampere\per\square\second}}
}
\newglossaryentry{larmor}{parent=physics,name=\ensuremath{\rho},
type=symbols,
sort={Larmor radius},
description={Larmor radius},
symbol={\si{\meter}}
}
\newglossaryentry{plasmaphysics}{name={Plasma physics quantitites},
type=symbols,
description={\nopostdesc},
symbol={},
}
\newglossaryentry{vperp}{parent=plasmaphysics,name=\ensuremath{v_{\perp}},
type=symbols,
sort=velocity,
description={Field-perpendicular velocity},
symbol={\si{\meter\per\second}}
}
\newglossaryentry{flowvelocity}{parent=plasmaphysics,name=\ensuremath{\vect{v_{\text{f}}}},
type=symbols,
sort=velocity,
description={Plasma flow velocity},
symbol={\si{\meter\per\second}}
}
\newglossaryentry{p}{parent=plasmaphysics,name=\ensuremath{p},
type=symbols,
sort=pressure,
description={Plasma pressure},
symbol={\si{\kg\per\meter\per\square\second}}
}
\newglossaryentry{V}{parent=plasmaphysics,name=\ensuremath{V},
type=symbols,
sort=volume,
description={Plasma volume},
symbol={\si{\cubic\meter}}
}
\newglossaryentry{viscosity}{parent=plasmaphysics,name=\ensuremath{\matr{\pi}},
type=symbols,
sort=viscosity,
description={Plasma viscosity tensor},
symbol={\si{\kg\per\meter\per\second}}
}
\newglossaryentry{Te}{parent=plasmaphysics,name=\ensuremath{T_{\mathrm{e}}},
type=symbols,
sort=temperature,
description={Electron temperature},
symbol={\si{\eV}}
}
\newglossaryentry{Ti}{parent=plasmaphysics,name=\ensuremath{T_{\mathrm{i}}},
type=symbols,
sort=temperature,
description={Ion temperature},
symbol={\si{\eV}}
}
\newglossaryentry{Tn}{parent=plasmaphysics,name=\ensuremath{T_{\mathrm{n}}},
type=symbols,
sort=temperature,
description={Neutral temperature},
symbol={\si{\eV}}
}
\newglossaryentry{Trad}{parent=plasmaphysics,name=\ensuremath{T_{\mathrm{rad}}},
type=symbols,
sort=temperature,
description={Radiation temperature},
symbol={\si{\eV}}
}
\newglossaryentry{ni}{parent=plasmaphysics,name=\ensuremath{n_{\mathrm{i}}},
type=symbols,
sort=density,
description={Ion density},
symbol={\si{\per\cubic\meter}}
}
\newglossaryentry{ne}{parent=plasmaphysics,name=\ensuremath{n_{\mathrm{e}}},
type=symbols,
sort=density,
description={Electron density},
symbol={\si{\per\cubic\meter}}
}
\newglossaryentry{nn}{parent=plasmaphysics,name=\ensuremath{n_{\mathrm{n}}},
type=symbols,
sort=density,
description={Neutral density},
symbol={\si{\per\cubic\meter}}
}
\newglossaryentry{nimp}{parent=plasmaphysics,name=\ensuremath{n_{\mathrm{imp}}},
type=symbols,
sort=density,
description={Impurity density},
symbol={\si{\per\cubic\meter}}
}
\newglossaryentry{te}{parent=plasmaphysics,name=\ensuremath{\tau_{\mathrm{E}}},
type=symbols,
sort=time,
description={Energy confinement time},
symbol={\si{\second}}
}
\newglossaryentry{radialShift}{parent=plasmaphysics,name=\ensuremath{\makeup{\Delta} R},
type=symbols,
sort=confinement,
description={Radial shift of the magnetic axis},
symbol={\si{\meter}}
}
\newglossaryentry{plasmaBeta}{parent=plasmaphysics,name=\ensuremath{\beta},
type=symbols,
sort=confinement,
description={Plasma beta},
symbol={}}
\newglossaryentry{Itor}{parent=plasmaphysics,name=\ensuremath{I_{\mathrm{tor}}},
type=symbols,
sort=confinement,
description={Toroidal plasma current},
symbol={\si{\ampere}}
}
\newglossaryentry{Ibs}{parent=plasmaphysics,name=\ensuremath{I_{\mathrm{bs}}},
type=symbols,
sort=confinement,
description={Bootstrap current},
symbol={\si{\ampere}}
}
\newglossaryentry{Ips}{parent=plasmaphysics,name=\ensuremath{I_{\mathrm{ps}}},
type=symbols,
sort=confinement,
description={Pfirsch-Schlueter current},
symbol={\si{\ampere}}
}
\newcommand{\quer}[1]{\mathrel{\hbox{-}\mkern-6.55mu #1}}
\newglossaryentry{ibar}{parent=plasmaphysics,name=\ensuremath{\quer{\iota}},
type=symbols,
sort=confinement,
description={Rotational transform},
symbol={}}
\newglossaryentry{shear}{parent=plasmaphysics,name=\ensuremath{s},
type=symbols,
sort=confinement,
description={Shear,radial derivative of \ensuremath{\quer{\iota}}},
symbol={}}
\newglossaryentry{ibarCF}{parent=plasmaphysics,name=\ensuremath{\quer{\iota}_{\mathrm{CF}}},
type=symbols,
sort=confinement,
description={Current free rotational transform},
symbol={}}
\newglossaryentry{Vp}{parent=plasmaphysics,name=\ensuremath{V_\mathrm{p}},
type=symbols,
sort=potential,
description={Plasma potential},
symbol={\si{\volt}}
}
\newglossaryentry{Vf}{parent=plasmaphysics,name=\ensuremath{V_\mathrm{f}},
type=symbols,
sort=potential,
description={Floating potential},
symbol={\si{\volt}}
}
\newglossaryentry{Isat}{parent=plasmaphysics,name=\ensuremath{I_\text{sat}},
type=symbols,
sort=current,
description={Ion saturation current},
symbol={\si{\ampere}}
}
\newglossaryentry{esat}{parent=plasmaphysics,name=\ensuremath{I_{e,\text{sat}}},
type=symbols,
sort=current,
description={Electron saturation current},
symbol={\si{\ampere}}
}
\newglossaryentry{jsat}{parent=plasmaphysics,name=\ensuremath{j_\text{sat}},
type=symbols,
sort=current,
description={Ion saturation current density},
symbol={\si{\ampere\per\square\meter}}
}
\newglossaryentry{csound}{parent=plasmaphysics,name=\ensuremath{c_\mathrm{s}},
type=symbols,
sort=velocity,
description={Ion sound speed},
symbol={\si{\meter\per\second}}
}
\newglossaryentry{Vbias}{parent=plasmaphysics,name=\ensuremath{V_\mathrm{bias}},
type=symbols,
sort=potential,
description={Bias voltage},
symbol={\si{\volt}}
}
\newglossaryentry{r}{parent=plasmaphysics,name=\ensuremath{r},
type=symbols,
sort=radius,
description={Minor radius},
symbol={\si{\meter}}
}
\newglossaryentry{ra}{parent=plasmaphysics,name=\ensuremath{r_a},
type=symbols,
sort=radius,
description={Minor radius of the last closed flux surface},
symbol={\si{\meter}}
}
\newglossaryentry{reff}{parent=plasmaphysics,name=\ensuremath{r_{\mathrm{eff}}},
type=symbols,
sort=radius,
description={Effective minor radius},
symbol={\si{\meter}}
}
\newglossaryentry{R}{parent=plasmaphysics,name=\ensuremath{R},
type=symbols,
sort=radius,
description={Major radius},
symbol={\si{\meter}}
}
\newglossaryentry{normminrad}{parent=plasmaphysics,name=\ensuremath{\varrho},
type=symbols,
sort=radius,
description={Normalised minor radius},
symbol={}}
\newglossaryentry{aspectRatio}{parent=plasmaphysics,name=\ensuremath{\epsilon},
type=symbols,
sort=radius,
description={Aspect ratio},
symbol={\si{}}
}
\newglossaryentry{Nfp}{parent=plasmaphysics,name=\ensuremath{N_{\text{fp}}},
type=symbols,
sort=number,
description={Number of field periods},
symbol={\si{}},
}
\newglossaryentry{phiEdge}{parent=plasmaphysics,name=\ensuremath{\phi_{\mathrm{edge}}},
type=symbols,
sort=magnetic flux,
description={Total enclosed magnetic toroidal flux},
symbol={\si{\volt\second}}
}
\newglossaryentry{Pconv}{parent=plasmaphysics,name=\ensuremath{P_{\mathrm{conv}}},
type=symbols,
sort=power,
description={Convective power},
symbol={\si{\watt}}
}
\newglossaryentry{Prad}{parent=plasmaphysics,name=\ensuremath{P_{\mathrm{rad}}},
type=symbols,
sort=power,
description={Radiated power},
symbol={\si{\watt}}
}
\newglossaryentry{Pheat}{parent=plasmaphysics,name=\ensuremath{P_{\mathrm{heat}}},
type=symbols,
sort=power,
description={Total heating power},
symbol={\si{\watt}}
}
\newglossaryentry{Lc}{parent=plasmaphysics,name=\ensuremath{L_{\mathrm{c}}},
type=symbols,
sort=length,
description={Connection length},
symbol={\si{\meter}}
}
\newglossaryentry{heatflux}{parent=plasmaphysics,name=\ensuremath{q},
type=symbols,
sort=power flux,
description={Heat flux or heat load},
symbol={\si{\MW\per\square\meter}}
}
\newglossaryentry{Zav}{parent=plasmaphysics,name=\ensuremath{Z_{\mathrm{av}}},
type=symbols,
sort=ion charge,
description={Average ion charge},
symbol={\si{}},
}
\newglossaryentry{Zeff}{parent=plasmaphysics,name=\ensuremath{Z_{\mathrm{eff}}},
type=symbols,
sort=ion charge,
description={Effective ion charge},
symbol={\si{}},
}
\newglossaryentry{iongyro}{parent=plasmaphysics,name=\ensuremath{\rho_{\text{i}}},
type=symbols,
sort=radius,
description={Ion gyro (Lamor) radius},
symbol={\si{\meter}},
}
\newglossaryentry{mi}{parent=plasmaphysics,name=\ensuremath{m_{\text{i}}},
type=symbols,
sort=mass,
description={Ion mass},
symbol={\si{\kilo\gram}},
}
\newglossaryentry{ndl}{parent=plasmaphysics,name=\ensuremath{n\text{d}\ell},
type=symbols,
sort=density,
description={Line integrated density},
symbol={\si{\per\square\meter}},
}
\newglossaryentry{Wdia}{parent=plasmaphysics,name=\ensuremath{W_\text{dia}},
type=symbols,
sort=energy,
description={Diamagnetic energy},
symbol={\si{\joule}},
}
\newglossaryentry{Qsci}{parent=plasmaphysics,name=\ensuremath{Q_{\text{sci}}},
type=symbols,
sort=gain,
description={Fusion energy scientific gain factor},
symbol={\si{}}
}
\newglossaryentry{Qeng}{parent=plasmaphysics,name=\ensuremath{Q_{\text{eng}}},
type=symbols,
sort=gain,
description={Fusion energy engineering gain factor},
symbol={\si{}}
}
\newglossaryentry{epseff}{parent=plasmaphysics,name=\ensuremath{\epsilon_\text{eff}},
type=symbols,
sort=effective ripple,
description={Effective ripple},
symbol={\si{}},
}
\newglossaryentry{Halpha}{parent=plasmaphysics,name=\ensuremath{\text{H}_{α}\xspace},
type=symbols,
sort=Wavelengths,
description={Hydrogen $\alpha$ transition line,Balmer series transition $n=3 \rightarrow n=2$,\SI{656.5}{\nano\meter}},
symbol=\si{\nano\meter},
}
\newglossaryentry{sxb}{parent=plasmaphysics,name=\ensuremath{\text{S/XB}},
type=symbols,
sort=Coefficients,
description={Ratio of ionisation,excitation and branching ratio. Inverse photons per neutral.},
symbol=,
}
\newglossaryentry{particleFlux}{parent=plasmaphysics,name=\ensuremath{\Gamma},
type=symbols,
sort=Particle flux,
description={Particle flux},
symbol=\si{\per\second\per\square\meter},
}
\newglossaryentry{stf}{parent=plasmaphysics,name=\ensuremath{\gamma_{\text{s}}},
type=symbols,
sort=Sheath transmission coefficient,
description={Sheath transmission coefficient},
symbol={\si{}},
}
\newglossaryentry{frad}{parent=plasmaphysics,name=\ensuremath{f_{\text{rad}}},
type=symbols,
sort=Fraction,
description={Radiated power fraction},
symbol={\si{}},
}
\newglossaryentry{mfpi}{parent=plasmaphysics,name=\ensuremath{\lambda_\text{mfp,i}},
type=symbols,
sort=Length,
description={Mean free path length of ionisation},
symbol={\si{\meter}},
}
\newglossaryentry{frec}{parent=plasmaphysics,name=\ensuremath{f_\text{rec}},
type=symbols,
sort=Fraction,
description={Fraction of ions recycled as neutrals at PFCs},
symbol={\si{}},
}
\newglossaryentry{Pecrh}{parent=plasmaphysics,name=\ensuremath{P_\text{ECRH}},
type=symbols,
sort=Power,
description={ECRH Power},
symbol={\si{\watt}},
}

\newacronym{machineLearning}{Machine Learning}{\nopostdesc}\newacronym[parent=machineLearning]{ML}{ML}{machine learning}\newacronym[parent=machineLearning]{RL}{RL}{reinforcement learning}\newacronym[parent=machineLearning,plural=NNs,firstplural=artificial neural networks (NNs)]{NN}{NN}{artificial neural network}\newacronym[parent=machineLearning]{EI}{EI}{expected improvement}\newacronym[parent=machineLearning,plural=FF-FCs,firstplural=feed-forward fully-connected neural networks (FF-FC)]{FFFC}{FF-FC}{feed-forward fully-connected neural network}\newacronym[parent=machineLearning,plural=MLPs,firstplural=multilayer perceptrons (MLP)]{MLP}{MLP}{multilayer perceptron}\newacronym[parent=machineLearning,plural=CNNs,firstplural=convolutional neural networks (CNNs)]{CNN}{CNN}{convolutional neural network}\newacronym[parent=machineLearning,plural=DCNNs,firstplural=deep convolutional neural networks (DCNNs)]{DCNN}{DCNN}{deep convolutional neural network}\newacronym[parent=machineLearning,plural=IRNNs,firstplural=Inception-ResNets (IRNNs)]{IRNN}{IRNN}{Inception-ResNet}\newacronym[parent=machineLearning,plural=DINNs,firstplural=deep inception neural networks (DINNs)]{DINN}{DINN}{deep inception neural network}\newacronym[parent=machineLearning,plural=GANs,firstplural=generative adversarial neural networks (GANs)]{GAN}{GAN}{generative adversarial neural network}\newacronym[parent=machineLearning,plural=DQNs,firstplural=deep Q-networks]{DQN}{DQN}{deep Q-network}\newacronym[parent=machineLearning,plural=AEs,firstplural=autoencoders]{AE}{AE}{autoencoder}\newacronym[parent=machineLearning,plural=VAEs,firstplural=variational autoencoders]{VAE}{VAE}{variational autoencoder}\newacronym[parent=machineLearning,plural=PINNs,firstplural=physics informed neural networks]{PINN}{PINN}{physics informed neural network}\newacronym[parent=machineLearning,plural=DeepONets,firstplural=deep operator networks]{DeepONet}{DeepONet}{deep operator network}\newacronym[parent=machineLearning]{MNIST}{MNIST}{MNIST}\newacronym[parent=machineLearning]{ReLU}{ReLU\xspace}{rectified linear unit}\newacronym[parent=machineLearning]{SeLU}{SeLU\xspace}{scaled exponential linear unit}\newacronym[parent=machineLearning]{SiLU}{SiLU\xspace}{sigmoid linear unit}\newacronym[parent=machineLearning]{rmse}{rmse\xspace}{root-mean-square error}\newacronym[parent=machineLearning]{mse}{mse\xspace}{mean-squared error}\newacronym[parent=machineLearning]{msd}{msd\xspace}{mean-squared deviation}\newacronym[parent=machineLearning]{nmse}{nmse\xspace}{normalized mean-squared error}\newacronym[parent=machineLearning]{rmsd}{rmsd\xspace}{root-mean-squared deviation}\newacronym[parent=machineLearning]{nrmsd}{nrmsd\xspace}{normalized root-mean-squared deviation}\newacronym[parent=machineLearning]{nrmse}{nrmse\xspace}{normalized root-mean-squared error}\newacronym[parent=machineLearning]{mae}{mae\xspace}{mean absolute error}\newacronym[parent=machineLearning]{mape}{mape\xspace}{mean absolute percentage error}\newacronym[parent=machineLearning]{TPE}{TPE}{tree-structured Parzen estimator}\newacronym[parent=machineLearning]{NAS}{NAS}{neural architecture search}\newacronym[parent=machineLearning]{NES}{NES}{neural ensemble search}\newacronym[parent=machineLearning]{SMBO}{SMBO}{sequential model-based optimization}\newacronym[parent=machineLearning]{BFGS}{BFGS}{Broyden–Fletcher–Goldfarb–Shanno}\newacronym[parent=machineLearning]{LBFGS}{L-BFGS}{limited-memory Broyden–Fletcher–Goldfarb–Shanno}\newacronym[parent=machineLearning]{AD}{AD}{automatic differentiation}
\newglossaryentry{ml}{name={Machine learning quantitites},
type=symbols,
description={\nopostdesc},
symbol={},
}
\newglossaryentry{lossfunction}{name=\ensuremath{\mathcal{L}},
type=symbols,
parent=ml,
sort=loss,
description={Loss function},
symbol={\si{}}
}
\everymath=\expandafter{\the\everymath\displaystyle}\usepackage{pgf}\usepackage{import}\usepackage{afterpage}\usepackage{graphics}\usepackage{mathtools}\usepackage{amsbsy}\usepackage{mathrsfs}\makeatletter\@ifpackageloaded{underscore}{}{\usepackage[strings]{underscore}}\makeatother
\usetikzlibrary{external}
\begin{document}\typeout{----- BEGIN DOCUMENT -----}\typeout{}\typeout{--------------------------------}\typeout{----- Document properties: -----}\typeout{--------------------------------}\typeout{Font size: \the\fsize}\typeout{Text width: \the\textwidth}\typeout{--------------------------------}\typeout{--------------------------------}\ifthenelseproperty{compilation}{titlepage}{\ifKOMA
\pagestyle{empty}\title{\getproperty{document}{title}\ifpropertydefined{document}{status}{\thanks{ This is the Accepted Manuscript version of an article accepted for publication in \getproperty{document}{journal}. \getproperty{document}{editor} is not responsible for any errors or omissions in this version of the manuscript or any version derived from it. This Accepted Manuscript is published under a \getproperty{document}{license} licence. The Version of Record is available online at \url{\getproperty{document}{doi}}
}
}}
\author[\getproperty{author}{affiliationindices}]{\getproperty{author}{firstname} \getproperty{author}{familyname}}
\affil[1]{\getproperty{affiliations}{1}}
\ifpropertydefined{coauthor1}{familyname}{\author[\getproperty{coauthor1}{affiliationindices}]{\getproperty{coauthor1}{firstname} \getproperty{coauthor1}{familyname}}}{}\ifpropertydefined{coauthor2}{familyname}{\author[\getproperty{coauthor2}{affiliationindices}]{\getproperty{coauthor2}{firstname} \getproperty{coauthor2}{familyname}}}{}\ifpropertydefined{coauthor3}{familyname}{\author[\getproperty{coauthor3}{affiliationindices}]{\getproperty{coauthor3}{firstname} \getproperty{coauthor3}{familyname}}}{}\ifpropertydefined{coauthor4}{familyname}{\author[\getproperty{coauthor4}{affiliationindices}]{\getproperty{coauthor4}{firstname} \getproperty{coauthor4}{familyname}}}{}\ifpropertydefined{coauthor5}{familyname}{\author[\getproperty{coauthor5}{affiliationindices}]{\getproperty{coauthor5}{firstname} \getproperty{coauthor5}{familyname}}}{}\ifpropertydefined{coauthor6}{familyname}{\author[\getproperty{coauthor6}{affiliationindices}]{\getproperty{coauthor6}{firstname} \getproperty{coauthor6}{familyname}}}{}\ifpropertydefined{coauthor7}{familyname}{\author[\getproperty{coauthor7}{affiliationindices}]{\getproperty{coauthor7}{firstname} \getproperty{coauthor7}{familyname}}}{}\ifpropertydefined{coauthor8}{familyname}{\author[\getproperty{coauthor8}{affiliationindices}]{\getproperty{coauthor8}{firstname} \getproperty{coauthor8}{familyname}}}{}\ifpropertydefined{coauthor9}{familyname}{\author[\getproperty{coauthor9}{affiliationindices}]{\getproperty{coauthor9}{firstname} \getproperty{coauthor9}{familyname}}}{}\ifpropertydefined{coauthor10}{familyname}{\author[\getproperty{coauthor10}{affiliationindices}]{\getproperty{coauthor10}{firstname} \getproperty{coauthor10}{familyname}}}{}\ifpropertydefined{coauthor11}{familyname}{\author[\getproperty{coauthor11}{affiliationindices}]{\getproperty{coauthor11}{firstname} \getproperty{coauthor11}{familyname}}}{}\ifpropertydefined{coauthors}{W7Xteaminclude}{\author[ ]{the W7-X team}}{}\ifpropertydefined{affiliations}{2}{\small \affil[2]{\getproperty{affiliations}{2}}
}{}\ifpropertydefined{affiliations}{3}{\affil[3]{\getproperty{affiliations}{3}}
}{}\ifpropertydefined{affiliations}{4}{\affil[4]{\getproperty{affiliations}{4}}
}{}\ifpropertydefined{affiliations}{5}{\affil[5]{\getproperty{affiliations}{5}}
}{}\ifpropertydefined{affiliations}{*}{\affil[*]{\getproperty{affiliations}{*}}
}{}\date{}\maketitle
\pagestyle{scrheadings}\else
\pagestyle{empty}\title{\getproperty{document}{title}}
\author{\getproperty{author}{firstname} \getproperty{author}{familyname}}
\date{\getproperty{document}{date}}
\maketitle
\fi
}{}\ifthenelseproperty{compilation}{abstract}{\ifthenelseproperty{compilation}{clsdefineschapter}{\ifKOMA \addchap[Abstract]{Abstract}\else
\chapter[Abstract]{Abstract}\fi
}{\ifKOMA \addsec[Abstract]{Abstract}\else
\section[Abstract]{Abstract}\fi
}
We present a novel approach to compute three-dimensional magnetohydrodynamic equilibria with isotropic pressure profiles and nested surfaces by parametrizing Fourier modes with artificial neural networks.
The full nonlinear global force residual of single equilibria across the volume in real space is then minimized with first order optimizers and compared to equilibria computed by conventional solvers.
Already,we observe competitive computational cost to arrive at the same minimum residuals computable with existing codes.
With increased computational cost,lower minima of the residual are \replaced[id=TT]{computable}{computed} with the neural networks \added[id=TT]{than with any other tested solver},establishing a new lower bound for the force residual.
We use minimally complex neural networks,and we expect significant improvements for solving not only single equilibria with neural networks,but also for creating neural network models valid over continuous distributions of equilibria.
}{}\ifthenelseproperty{compilation}{glossaries}{\glsresetall }{}\ifthenelseproperty{compilation}{toc}{\input{templates_Default_toc.tex}}{}\newpage
\pagestyle{empty}\pagestyle{fancy}\section{Introduction}\label{sec:introduction}\Gls{MHD} models describe experimental and astrophysical plasmas as single species fluids with electromagnetic effects in the long-wavelength and low-frequency limit.
By considering the fluid as a perfect conductor without electric resistance,this system of \glspl{PDE} reduces to the \emph{ideal} \gls{MHD} equations.
The equilibrium states of ideal \gls{MHD} are a baseline in the description of magnetically confined,ring-shaped plasmas in fusion experiments.
One goal of such experiments is the achievement of self-sustaining hot plasmas in a torus that are amenable to the release of excess energy by fusing light nuclei.\\
Computation of an equilibrium magnetic field is a first step for many subsequent calculations in axisymmetric and non-axisymmetric plasmas.
Assuming the magnetic field to be axisymmetric reduces the ideal \gls{MHD} equations to the Grad-Shafranov equation,but this assumption can be too strong for control applications of axisymmetric tokamaks~\cite{Fasoli2016,Ham2015}.
Analytical models for the 3D ideal \gls{MHD} equations exist for high-pressure and large-aspect configurations~\cite{Freidberg2014},low-order Fourier modes of the magnetic field in a subset of reactor-relevant stellarators~\cite{Nikulsin2024} and for the description of equilibrium magnetic fields close to axis~\cite{Jorge2020}.
Still,global analytic solutions of ideal \gls{MHD} equilibria with finite pressure in tori are missing~\cite{Kaiser1994},which motivated the creation of many numerical solvers for 3D plasma equilibria~\cite{Hirshman,Dudt2020,Hindenlang2025,hudson2012,Reiman1988,Suzuki2006}.
The optimization space of non-axisymmetric fields is orders of magnitude larger than that of axisymmetric devices~\cite{Helander2014},and certain geometries only become computable with novel solvers,for example,figure-8 stellarators~\cite{Plunk2025}.
An assumption that many solvers implement,especially in optimization workflows,is the existence of nested magnetic \emph{flux surfaces}.
With this assumption,magnetic islands and chaotic regions are excluded in the description of the plasma,and this simplifies computation by ensuring integrability over the whole plasma volume.
Flux surfaces are isobaric surfaces on which the toroidal magnetic flux $2\pi \psi$ and poloidal magnetic flux $2\pi \chi$ are constant~\cite{Helander2014}.
The existence of flux surfaces in three-dimensional plasma boundaries is proven for stepped pressure profiles including island regions close to axisymmetry~\cite{Bruno1996},but not for strongly non-axisymmetric cases with continuous pressure profiles.\\
Fast and accurate computational models for 3D ideal \gls{MHD} equilibria can enable real-time control of plasmas,flight-simulators and other data-intensive applications.
To evaluate \glspl{NN} as a parametrization for Fourier-decomposed,3D ideal \gls{MHD} equilibria,we examine the lower bound on \gls{NN} complexity.
Specifically,we investigate the performance of two-hidden-layer \glspl{MLP} and their hidden layer widths to represent Fourier modes of magnetic fields of single fixed-boundary equilibria with isotropic pressure profiles.
\subsection{Motivation}\label{sec:motivation}Fast,accurate and continuous models of three-dimensional plasma equilibrium evolution expedite data analysis pipelines~\cite{Merlo2023},enhance real-time control~\cite{Fasoli2016} and enable a zoo of other data-intensive applications for stellarators and tokamaks.
Stellarators with inherently stable plasmas rely much less on active control than tokamaks,but control of transport or turbulence in stellarators will likely benefit from rapid ideal \gls{MHD} equilibrium information.
Significantly reducing \replaced[id=TT]{inference time of a single equilibrium}{the runtime for individual equilibrium solves} is of great interest to real-time interpretation of diagnostic data~\cite{Merz2013}.
This work explores the minimum size of \gls{MLP} required for solving single equilibria.
The results of the presented investigation (see figure~\ref{fig:scan_mn_w7xM10N10} and~\ref{fig:scan_time_w7xM10N10}) are a valuable starting-point to construct continuous operator models that support rapid inference.
Currently existing stellarator equilibrium solvers are not able to provide equilibria in less than $\mathcal{O}(10^{-3}\si{s})$ without massive parallelization,driving up operational costs for real-time control systems.
Creating a look-up table of equilibria for the configuration space of a specific device using a conventional solver will incur some interpolation error.
\glspl{NN}  provide continuous output functions and,trained on \glspl{PDE} or diagnostic data,can achieve super-resolution and reveal hidden relations between diagnostics~\cite{Rossi2025,Jalalvand2024}.\newline
\textcite{Merlo2023} shows benefit to stellarator optimization and acceleration of bayesian inference by several orders of magnitude using an operator model trained on the configuration space of \gls{W7X} as calculated by the ideal \gls{MHD} solver \texttt{VMEC}.\newline
This work tries to answer if it is possible to skip conventional solvers altogether,not creating any datasets and thus relying only on the physics of the equilibrium problem to train \gls{NN}.
Training a single \gls{NN} model on a single equilibrium is a first step in that direction.
We extrapolate from the results of~\textcite{VanMilligen1995},where a \gls{MLP} with one hidden layer was sufficient to solve stellarator equilibria,and investigate minimally complex \gls{MLP} with two hidden layers trained solely on the ideal \gls{MHD} force residual. Furthermore,~\textcite{Paluzo-Hidalgo2020} prove that maps between triangulable spaces,which the authors write include most real-world problems,can be solved with two-hidden-layer \glspl{NN}.\newline
Like the modern ideal \gls{MHD} solver \texttt{DESC},the code implementing this work utilizes the automatic differentiation library \texttt{JAX},which offers computation along a graph with tangent spaces,enabling the calculation of first order derivatives with a small computational and memory overhead~\cite{jax2018github}.
Higher order derivative computation incurs exponentially larger resource requirements and the presented approach requires computation of second order derivatives because we optimize for the strong form of the ideal \gls{MHD} residual.
This automatic differentiation saves compute time and provides more precise gradients compared to finite differencing gradient computation schemes in larger stellarator optimization workflows~\cite{Dudt2022}.\newline
\glspl{NN} applied to solve physics equations in nuclear fusion research is promising~\cite{Seo2024},and this work introduces a bridge connecting \glspl{NN} with the ideal \gls{MHD} equilibrium problem.
The modular approach of this work allows for simple investigation of novel \gls{NN} architectures,training schemes and optimizers,and also the addition of new physics.
\section{Related Work}\label{sec:related_work}\subsection{3D ideal MHD equilibrium}\label{subsec:3d-magnetohydrodynamic-equilibrium}Ideal \gls{MHD} equilibria are computed by minimizing either the weak or strong form of the ideal,stationary \gls{MHD} equations in a map between two different coordinate systems.
One of these coordinate systems is chosen such that the magnetic field lines are straight,namely the magnetic coordinates,and the other is the cartesian or cylindrical coordinate system.
This work uses the inverse formulation,that is the \glspl{MLP} map from magnetic back to cylindrical coordinates $\pmb{\alpha}=[s,\theta,\zeta]^\mathsf{T}  \rightarrow [R,\lambda,Z]^\mathsf{T}$,to minimize residuals of finite-pressure,fixed-boundary ideal \gls{MHD} equilibria evaluated in the cartesian frame.\newline
Stellarator equilibria are often computed with toroidal symmetry with $N_\mathrm{FP}$ equal toroidal segments,each segment,or field period,occupying $2\pi/N_\mathrm{FP}$ of the full torus.
In the magnetic coordinates,the normalized flux $s=\psi/\psi_\mathrm{b} \in [0,1]$ can act as the radial coordinate with $\psi_\mathrm{b}$ denoting the toroidal flux enclosed by the plasma boundary,$\theta \in [0,2\pi]$ is a poloidal angle in geometric coordinates and $\zeta=\phi \in [0,2\pi/N_\mathrm{FP}]$ is a toroidal angle equal to the geometric cylindrical angle.
$R$ and $Z$ are part of the cylindrical coordinate system with its origin coinciding with the major axis of the plasma.
In the inverse approach to the ideal \gls{MHD} equilibrium problem,the magnetic field is parametrized by some continuous radial coordinate,which imposes a continuous distribution of flux surfaces,yielding a nested set of flux surfaces.
It follows that no magnetic islands or chaotic regions can be expressed in the inverse approach,however,this assumption simplifies the computation compared to the direct map $[R,\lambda,Z]^\mathsf{T} \rightarrow  [s,\theta,\zeta]^\mathsf{T}$.\newline
It is common practice to decompose the poloidal and toroidal angles in a 2D Fourier series with angle $\phifmn=m\,\theta - n\,N_\mathrm{FP}\,\zeta$.
Fourier expanding the independent variables reduces the input dimensionality of the ideal \gls{MHD} equilibrium problem by two: $s \rightarrow (R,\lambda,Z)$~\cite{Lao1981},but comes with the cost associated with finite truncation of the chosen basis.
The subscript $m$ denotes poloidal modes as $m=\{0,1,2,\cdots,M-1\}$ for a total number of $M$ poloidal modes and the subscript $n$ denotes toroidal modes  $n=\{-N,-N+1,\cdots,0,\cdots,N-1,N\}$ for a total number of $2N+1$ toroidal modes.
The dependent variables are then split into cosine and sine components,denoted by superscripts $c$ and $s$ respectively (see equations~\eqref{eqn:R_NN_pred}-\eqref{eqn:Z_NN_pred}).
Real valued coordinates $X\in \{R,\lambda,Z\}$ can then be recovered via the series
\begin{equation}X=\sum_{m=0,n=-N}^{M-1,N} X_{mn}^\mathrm{c} \,\cos(\phifmn) + X_{mn}^\mathrm{s} \,\sin(\phifmn)
\label{eqn:basic_fourier}\end{equation}with $X_{mn}=\overline{X_{-m-n}}$ such that $X_{m=0,n<0}$ and $X_{m<0,n}$ can be removed~\cite{Hirshman}.
The renormalizing stream function $\lambda$ and the $Z$ coordinate are periodic functions with zero mean,~i.e\ $\smallint \smallint X d\theta d\zeta=0$ for $X \in \{\lambda,Z\}$,which is analogous to setting the $(m,n)=(0,0)$ modes to $0$: $\lambda^s_{00}=Z^s_{00}=0$.
$\lambda$ converts the geometric poloidal angle $\theta$ to the magnetic poloidal angle $\theta^{\star}$: $\theta^{\star}=\theta+\lambda$ in which field lines are straight and also facilitates numerical convergence by condensing shape information of the flux surfaces into the lower modes of the Fourier series expansion~\cite{Hirshman}.\newline
The exact definition of the poloidal angle $\theta$ can be set freely as long as it is periodic,and the Jacobian of the transformation between the two coordinate systems stays finite and does not change sign.
Lastly,stellarator equilibria are often assumed to be up-down symmetric in at least two toroidal angles per field period,expressed as
\begin{align}R(s,\theta,\zeta) &=R(s,-\theta,-\zeta)\nonumber\\
Z(s,\theta,\zeta) &=-Z(s,-\theta,-\zeta)
\end{align}If this \emph{stellarator symmetry}~\cite{Dewar1998} is assumed,the computational grid can be reduced by half in the poloidal coordinate $\theta\in[0,\pi]$ and the non-symmetric Fourier modes are omitted,which further reduces computational cost: $R_{mn}^\mathrm{s}=Z_{mn}^\mathrm{c}=\lambda_{mn}^\mathrm{c}=0$.
Ideal \gls{MHD} equilibria with nested flux surfaces,isotropic pressure and a fixed plasma boundary are then defined on the map between the magnetic and geometric coordinates and fully characterized by three invariants:
The radial pressure profile~$p(s)$,the rotational transform profile~$\iota(s)$ and the boundary geometry,in the following enforced in Fourier space as~$R_{mn}^\mathrm{c}(s=1)$ and~$Z_{mn}^\mathrm{s}(s=1)$~\cite{Kruskal1958}.
Replacement of the rotational transform profile with the toroidal current profile is also possible but not used in this work.
In the remainder of this section,the computation of the force residual is detailed:\newline
The covariant basis vectors of the inverse map are
$$
\mathbf{e}_s=\left[\begin{array}{c}\partial_s R \\
0            \\
\partial_s Z
\end{array}\right] \quad \mathbf{e}_{\theta}=\left[\begin{array}{c}\partial_{\theta} R \\
0                   \\
\partial_{\theta} Z
\end{array}\right] \quad \mathbf{e}_{\zeta}=\left[\begin{array}{c}\partial_{\zeta} R \\
R                  \\
\partial_{\zeta} Z
\end{array}\right]
\label{eqn:basis_vectors}$$
and their contravariants are $\mathbf{e}^i=\nabla i=\frac{\mathbf{e}_j \times \mathbf{e}_k}{\sqrt{g}}$ with $(i,j,k)$ a cyclic permutation of $\pmb{\alpha}$.
The partial derivative is denoted with $\partial \alpha_i X=\frac{\partial X}{\partial \alpha_i}$ for $X\in\{R,\lambda,Z\}$.
\begin{equation}\sqrt{g}=\mathbf{e}_s \cdot \mathbf{e}_\theta \times \mathbf{e}_\zeta=(\mathbf{e}^s \cdot \mathbf{e}^{\theta} \times \mathbf{e}^\zeta)^{-1}\label{eq:sqrt}\end{equation}is the Jacobian of the map.
Then,the metric tensor components can be easily computed~\cite{Hirshman}:
\begin{align}\hspace{-0.25cm}g_{ij} &=\mathbf{e}_i \mathbf{e}_j=\partial_i R \,\partial_j R + R^2 \,\partial_i \phi \,\partial_j \phi + \partial_i Z \,\partial_j Z\\
\hspace{-0.25cm}g^{ij} &=\mathbf{e}^i \mathbf{e}^j
\end{align}The force residual is the remainder of the ideal \gls{MHD} equations,which models the pressure balance between confining magnetic field and pressure exerted by the plasma.
In steady-state ($\partial_t=0$),the magnetic field has to counteract the plasma pressure for which the ideal \gls{MHD} equations reduce to
\begin{align}\mathbf{J} \times \mathbf{B} &=\nabla p \label{eqn:mhd_force}\\
\mu_\mathrm{0} \mathbf{J} &=\nabla \times \mathbf{B} \label{eqn:ampere}\\
\nabla \cdot \mathbf{B} &=0  \label{eqn:gauss}\end{align}with the vacuum permeability $\mu_\mathrm{0}$.
In the absence of magnetic islands,chaotic regions and plasma resistivity,the flux surfaces may be viewed as a continuous distribution of isobaric surfaces for $s\in[0,1]$,and the contravariant magnetic field can be expressed as~\cite{Helander2014}\begin{align}\label{eqn:b_f}\mathbf{B} &=\nabla \zeta \times \nabla \chi + \nabla \psi \times \nabla \theta^{\star}\\
&=B^s \mathbf{e}_s + B^\theta \mathbf{e}_\theta + B^\zeta \mathbf{e}_{\zeta}. \nonumber
\end{align}The assumption of nested flux surfaces for the magnetic topology is $B^s=\mathbf{B}\cdot \nabla s=0$.
The magnetic field~\eqref{eqn:b_f} together with Gauss's law~\eqref{eqn:gauss} results in the remaining magnetic field components
\begin{align}\mathbf{B}=\,\frac{\partial_s \psi}{\sqrt{g}} \left((\iota(s) \,- \,\partial_\zeta \lambda) \mathbf{e}_{\theta} +  (1+\partial_\theta \lambda) \mathbf{e}_{\zeta}\right).
\label{eqn:b_contra}\end{align}The magnetic field's covariant components are computed via
\begin{equation}B_i=B^{\theta} g_{i\theta} + B^{\zeta} g_{i\zeta}.
\label{eqn:b_cov}\end{equation}Using Ampere's law~\eqref{eqn:ampere},the contravariant current components are expressed as
\begin{align}J^i=\frac{1}{\mu_0 \sqrt{g}} \left( \partial_j B_k - \partial_k B_j \right)
\end{align}for cyclic permutations $(i,j,k)$ of $\pmb{\alpha}$.
The $\mathbf{B}$-field~\eqref{eqn:b_contra} inserted into the force balance equation~\eqref{eqn:mhd_force} reveals the two independent parts of the force residual~\cite{Hirshman}:
\begin{align}\mathbf{F} &=F_{s} \mathbf{e}^{s} + F_{h} \mathbf{e}^h\nonumber\\
&=(\nabla \times \mathbf{B}) \times \mathbf{B} - \mu_0 \nabla p
\label{eqn:ideal_mhd_force_balance}\end{align}with $\mathbf{e}^h=\sqrt{g}(B^\zeta \mathbf{e}^\theta - B^\theta \mathbf{e}^\zeta)=\nabla \theta - \iota \nabla \zeta$ denoting the helical direction.
Using the magnitude of this force vector
\begin{equation}||\mathbf{F}||_2=\sqrt {F_{s}^2 ||\mathbf{e}^s||_2 + F_{h}^2 ||\mathbf{e}^h||_2}
\label{eqn:f_magn}\end{equation}allows for a fair comparison between equilibrium solutions of different numerical solvers on equal computational grids.
This metric is then normalized with the gradient of the magnetic pressure $\langle |\nabla |B|^{2}/(2\mu_\mathrm{0})| \rangle_\mathrm{vol}$ with $\nabla |B|^2$ given by
\begin{align}\nabla |B|^2=2  (|B|  \nabla |B|)
\label{eqn:b_magnitude}\end{align}and $|B|=\sqrt{B\cdot B}$.
Another characteristic of the magnetic topology is the straight field line angle $\theta^{\star}$,which is also used in comparisons between solvers.
The volume average of some quantity $(\cdot)$ is denoted by $\langle (\cdot) \rangle_\mathrm{vol}$ and defined as
\begin{equation}\langle (\cdot) \rangle_\mathrm{vol}=\frac{1}{V} \int (\cdot) \sqrt{g} d\pmb{\alpha}\label{eqn:volume_average}\end{equation}with the plasma volume $V=\smallint \sqrt{g} \,d\pmb{\alpha}$.\\
Despite all the assumptions in this model,it is challenging to solve,partly due to the mixed elliptic-hyperbolic form of the equations and finite truncation of basis functions~\cite{Burby2023}.\newline
Many different solvers have been created,each with a unique parametrization and associated costs.
In the following,we will describe the numerical solvers \texttt{VMEC} and \texttt{DESC},focusing on the important aspects of each solver,introduce the \gls{NN}-based approach in more detail and finish with a comparison.
\subsection{VMEC }\label{sec:vmec}\texttt{VMEC}~\cite{Hirshman} is the workhorse of stellarator optimization~\cite{simsopt,stellopt},equilibrium reconstruction~\cite{stellopt,Hanson2009,Seal2017,Andreeva2019a} and inference of plasma parameters~\cite{stellopt,Bozhenkov2020,Beidler2021}.
Several experimental devices were constructed using \texttt{VMEC} in its free-boundary mode~\cite{Knowlton2005,Qian2023}.
Instead of the force residual~\eqref{eqn:f_magn},\texttt{VMEC} uses the weak formulation and minimizes the potential energy $W_\mathrm{pot}$ with a variational principle in magnetic coordinates
\begin{flalign}\label{eqn:wpot}W_\mathrm{pot}=\int \frac{|B|^2}{2\mu_\mathrm{0}} |\sqrt{g} \,| d\pmb{\alpha} + \int_0^1 \frac{M_\mathrm{m}(\rho)}{\gamma-1} (V')^{1-\gamma} d\rho&&
\end{flalign}with the specific heat ratio (or adiabatic index) $\gamma$ and some constant mass function $M_\mathrm{m}(\rho)$.
$V'(s)$ is the radial derivative of the volume $V(s)$ enclosed by the flux surface at $s$
\begin{align}V(s)=\int_0^s \int_0^{2\pi} \int_0^{2\pi} \sqrt{g} \,ds \,d\theta \,d\zeta.
\end{align}Variationally minimizing the potential energy reduces the computational load of minimization because first order gradients need to be calculated in the weak formulation only~\cite{Hirshman}.
Given stellarator symmetric equilibria,the dependent variables $R$,$\lambda$ and $Z$ are expanded in Fourier series \added[id=TT]{with poloidal and toroidal mode index $m$ and $n$}\begin{align}\hspace{-0.75cm} R(s,\theta,\zeta) &=\sum_{mn} R_{mn}^\mathrm{c} \cos(\phifmn)\label{eqn:R_fourier} \\[3pt]
\hspace{-0.75cm} \lambda(s,\theta,\zeta) &=\sum_{mn} \lambda_{mn}^\mathrm{s} \sin(\phifmn)\label{eqn:L_fourier} \\[3pt]
\hspace{-0.75cm} Z(s,\theta,\zeta) &=\sum_{mn} Z_{mn}^\mathrm{s} \sin(\phifmn).\label{eqn:Z_fourier}\end{align}Internally,\texttt{VMEC} uses a specialized double Fourier representation that is both efficient in terms of information in low modes and a unique representation.
This new representation fulfills the minimization of the following power spectrum
\begin{equation}\label{eqn:spectral_width}M_\mathrm{sp}=\sum_{mn} m^2 (R_{mn}^2 + Z_{mn}^2)
\end{equation}by construction~\cite{Hirshman1998}.
To further increase solution efficiency,\texttt{VMEC} uses pre-conditioning of its radial discretization~\cite{Hirshman1990} and calculates values on a \emph{half-grid},which only includes points $\theta \in [0,\pi]$ if the equilibrium possesses stellarator symmetry.
Last but not least,a common pattern in \texttt{VMEC} is the refinement of the radial grid,starting with a coarse discretization of the flux surfaces and reaching the final solution on a fine discretization—but this can lead to unphysical current spikes in the final solution at the coarse gridpoints~\cite{Panici2023}.
\texttt{VMEC} is a finite difference solver with associated drawbacks,for example,the requirement to interpolate values between flux surfaces and the introduction of numerical error due to finite discretization,and the magnetic axis calculated by \texttt{VMEC} lacks precision~\cite{Panici2023}.
Despite its minor drawbacks,\texttt{VMEC} is a viable candidate as a ground truth because of its empirical validation and widespread use in fusion research.
These drawbacks and the inability of current solvers to compute continuous models valid over continuous distributions of equilibria also motivated the presented work to inspect \glspl{NN} parametrizing \texttt{VMEC}'s Fourier basis (see equations~\eqref{eqn:R_fourier}-~\eqref{eqn:Z_fourier}) for single equilibria as a first step.
\subsection{DESC stellarator equilibrium solver}\label{sec:desc}The pseudo-spectral solver \texttt{DESC} uses a global,orthogonal basis set (on the unit disc) consisting of Zernike polynomials of the radial coordinate $\rho=\sqrt{s}$ and angle $\theta$ in conjunction with Fourier expanding  $\zeta$.
This basis avoids \texttt{VMEC}'s issues at the magnetic axis,reduces the number of parameters and is free from interpolation error between flux surfaces~\cite{Dudt2020}.
Replacing the finite differencing scheme of \texttt{VMEC} in radial direction with Zernike polynomials of order $L_\mathrm{ZP}$ provides a spectral parametrization in all three independent coordinates ($\rho,\theta,\zeta $).
For each $X \in \{R,\lambda,Z\}$ the Fourier Zernike basis is
\begin{flalign}X(\rho,\theta,\zeta)=\sum_{m=-M_\mathrm{D}}^{M_\mathrm{D}} \sum_{n=-N}^N \sum_{l=0}^{L_\mathrm{ZP}} X_{lmn} \mathscr{L}_l^m(\rho,\theta) \mathscr{F}^n(\zeta)&&
\label{eqn:desc_Fzernike}\end{flalign}where $M_\mathrm{D}$ is different from \texttt{VMEC}'s $M$ because it is the maximum poloidal mode number instead of the total number of modes~\cite{Panici2023}.
The Zernike polynomial defined on $\rho\in[0,1]$ and $\theta\in[0,2\pi]$ is
\begin{equation}\mathscr{L}_l^m(\rho,\theta)=
\begin{cases}\mathscr{R}_l^{|m|} \cos(|m|\theta) & \text{if } m \geq 0 \\
\mathscr{R}_l^{|m|} \sin(|m|\theta) & \text{if } m < 0
\end{cases}\label{eqn:desc_Zernike}\end{equation}with the Fourier decomposition defined for $\zeta \in[0,2\pi]$ as
\begin{equation}\mathscr{F}^n(\zeta)=
\begin{cases}\cos(|n|N_\mathrm{FP}\,\zeta) & \text{if } n \geq 0 \\
\sin(|n|N_\mathrm{FP}\,\zeta) & \text{if } n < 0
\end{cases}\label{eqn:desc_Fourier}\end{equation}The shifted Jacobi polynomial $\mathscr{R}_l^{|m|}$,defined for $m\geq 0$,is
\begin{equation}\mathscr{R}_l^{|m|} (\rho)=\sum_{s=0}^{(l-|m|)/2} \frac{(-1)^s(l-s)!}{ s!\left( \frac{l+|m|}{2} - s\right)! \left( \frac{l-|m|}{2} - s\right)!  }\hspace{0.1cm} \rho^{l-2s}.
\end{equation}This spectral representation fulfills the constraint on analytic functions~\eqref{eqn:analyticity} and enables analytical gradient computation of the geometric coordinates with respect to the magnetic coordinates~\cite{Dudt2020}.
Correct analytical representation of physical scalars at the axis is important for stability calculations~\cite{Panici2023}.
\texttt{DESC} can minimize the non-linear force error,or strong form,$\mathbf{F}=dW_\mathrm{pot}/d\pmb{\alpha}$ in real space,requiring computation and memory for second order derivatives,or the potential energy,i.e. the weak form,~\eqref{eqn:wpot} or many other compositions of targets in geometric or magnetic coordinates.
Quasi-Newton or least-squares optimizers implemented in \texttt{DESC} enable super-linear convergence in the system~\cite{Dudt2020}\begin{equation}\mathbf{f}(\mathbf{x}_\mathrm{sc},\mathbf{c}) \approx \mathbf{0}.
\label{eqn:desc_sysofeq}\end{equation}The boundary conditions $\mathbf{c}$,the spectral coefficients $\mathbf{x}_\mathrm{sc}$ and the residuals $\approx \mathbf{0}$ can be evaluated on one of many grids,for example,a quadrature grid which precisely integrates the chosen Fourier Zernike basis over the whole plasma volume.\newline
The optimization in \texttt{DESC} is carried out over parameters in a tangent space $\mathbf{y}$ with a particular solution $\mathbf{x}_\mathrm{p,sc}$,ensuring that all linear constraints are satisfied during minimization:
\begin{align}&\mathbf{A} \mathbf{x}_\mathrm{sc} &=\mathbf{c}\nonumber \\
\iff &\mathbf{x}_\mathrm{p,sc} + \mathbf{Z} \mathbf{y} &=\mathbf{c}\end{align}\texttt{DESC} can utilize continuation methods to speed up convergence and computes spectrally dense solutions in the sense of equation~\eqref{eqn:spectral_width}.
First,\texttt{DESC} solves an axisymmetric vacuum equilibrium which is then perturbed iteratively via two multipliers: $\eta_\mathrm{b} \in [0,1]$ for the toroidal boundary harmonics and~$\eta_\mathrm{p}\in [0,1]$ for the pressure coefficients until both arrive at $1$ and the final solution is obtained.
The perturbations are Taylor expansions of the loss functional~$\mathbf{f}(\mathbf{y} + \Delta \mathbf{y},\mathbf{c} + \Delta \mathbf{c})$ in some small parameter~$\epsilon$ to find the first-order direction $\mathbf{y}_1$~\cite{Conlin2022} that satisfies the perturbed constraints~$\mathbf{c} + \Delta \mathbf{c}$
\begin{align}\Delta \mathbf{y} &=\epsilon \mathbf{y}_1 + \epsilon^2 \mathbf{y}_2 + \cdots\nonumber\\
\Delta \mathbf{c} &=\epsilon \mathbf{c}_1 \\
0 &=\frac{\partial \mathbf{f}}{\partial \mathbf{y}}\epsilon \mathbf{y}_1 + \frac{\partial \mathbf{f}}{\partial \mathbf{c}}\epsilon \mathbf{c}_1\\[3pt]
\mathbf{y}_1 &=- \left( \frac{\partial \mathbf{f}}{\partial \mathbf{y}} \right)^{-1} \left( \frac{\partial \mathbf{f}}{\partial \mathbf{c}} c_1 \right).
\mathbf{y}_1 &=- \left( \frac{\partial \mathbf{f}}{\partial \mathbf{y}} \right)^{-1} \left( \frac{\partial \mathbf{f}}{\partial \mathbf{c}} c_1 \right).
\end{align}\texttt{DESC} uses concentric grids during minimization of the force residual by default,which are also used in this work.\newline
The flexibility of \texttt{DESC} and the computation of analytic gradients in the Fourier Zernike basis facilitated recent extensions to include broader stellarator optimization,e.g.~target functions for omnigeneous magnetic fields optimized with constraints~\cite{Conlin2024,Dudt2024}.
\newpage
\subsection{Neural Networks applied to 3D ideal MHD equilibrium problems}\label{sec:nn_to_3dMHD}One of the first publications about \glspl{PINN} used \gls{MHD}-based residuals in the domain of magnetically confined plasmas as loss functions to train \glspl{NN}~\cite{VanMilligen1995,VanMilligen1995a,Lagaris1998}.
Already then,~\textcite{VanMilligen1995a} describe that solutions using \glspl{MLP} need no finite-difference scheme,hinting at automatic differentiation.
Their approach requires a transformation from Fourier representation to real space and provides one of the first \glspl{PINN} to solve axisymmetric plasma equilibria.
In a follow-up publication,they show the initial feasibility of representing three-dimensional plasma equilibria with \glspl{NN} using $L2$-distance to data generated by \texttt{VMEC}~\cite{VanMilligen1995}.
These approaches were not pursued further until a decade later.
The first application of \glspl{NN} to equilibria of a real machine is presented by~\textcite{Sengupta2004}: A vacuum dataset in the space of the stellarator \gls{W7X} is used to train a \gls{NN} as a first step towards fast equilibrium reconstruction.
A follow-up publication deprecated the \glspl{NN} in favor of polynomials that showed similar performance to the \glspl{NN} models~\cite{Sengupta2007}.\newline
Driven by the availability of modern algorithms and hardware,~\textcite{Merlo2021} introduced a specialized operator \gls{NN} model that reproduces flux surface geometries over most of the operational space of \gls{W7X}.
This operator \gls{NN} was trained on a dataset of flux surface geometries,their first-order gradients as calculated by \texttt{VMEC} and a surrogate force residual which assumes $F_h=0$.
Progress made by~\textcite{Merlo2023} accelerated equilibrium reconstruction in the Bayesian framework \emph{Minerva}~\cite{Svensson2007,Svensson2010},which is used to interpret experimental data at \gls{W7X}.
Remarkably,the inclusion of a physics-based residual during training improves the accuracy of the operator model.
\added[id=TT]{Other neural surrogate models include models for axisymmetric equilibria computed with measured magnetic signals and the Grad-Shafranov residual in KSTAR~\cite{Joung2020},or models for core transport heat and particle fluxes in JET~\cite{VanDePlassche2020}.
The potential of \glspl{PINN} to improve transport simulations over classical finite difference methods is discussed in~\textcite{Seo2024a}.}\newline
\textcite{jang2024} first introduced operator \gls{PINN} for the direct map $NN: (R,Z,P,\epsilon,\kappa,\delta) \in \mathbb{R}^6 \mapsto \Psi(R,Z,P,\epsilon,\kappa,\delta) \in \mathbb{R}$ trained solely on physics targets over a space of axisymmetric tokamak equilibria.
Both $R$ and $Z$ coordinates are inputs of the \gls{NN},$P$ is a parameter for the pressure profile,$\epsilon$ is the inverse aspect ratio,the elongation is $\kappa$ and $\delta$ is the triangularity.
The \gls{NN}'s output is the total enclosed toroidal flux $\Psi$,which is sufficient to describe the direct map $(R,Z) \mapsto \Psi$ in axisymmetry.\newline
\begin{figure}\centering
\includegraphics[width=0.5\textwidth]{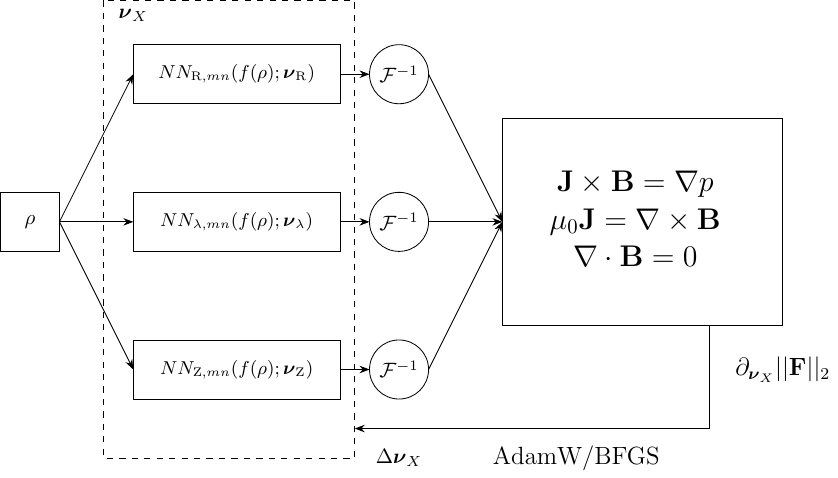}\caption{Illustration of the fully differentiable computational graph \added[id=TT]{of our contribution} with \glspl{NN} parameters $\pmb{\nu}_X$ and associated derivatives $\partial_{\pmb{\nu}_X} ||\mathbf{F}||_2$. The inverse Fourier transform is marked with $\mathcal{F}^{-1}$. \replaced[id=TT]{The strong form of the ideal \gls{MHD} residual requires up to second order gradients.}{Computing the \gls{MHD} force residual requires first and second order gradients of $(R,\lambda,Z)$ with respect to $\rho$.}}
\label{fig:comp_graph}\end{figure}The advances of~\cite{Merlo2023} and~\cite{Merlo2021},which used large models and large datasets,motivated this investigation of simple \glspl{NN} that parametrize Fourier modes to minimize the ideal \gls{MHD} equilibrium force residual.
Before training larger operator models,we want to take a step back and find a lower bound on the \gls{NN} complexity to parametrize solutions with low residual (see e.g. figures~\ref{fig:scan_mn_w7xM10N10} and~\ref{fig:scan_time_w7xM10N10}).
Furthermore,~\textcite{VanMilligen1995} showed that a single equilibrium problem can be solved by optimising parameters of a one-hidden-layer \gls{NN} from which we \replaced[id=TT]{continue}{extrapolate} and employ \gls{NN} with two-hidden layers to solve single equilibria.\newline
\replaced[id=TT]{The computational graph of our contribution is illustrated in figure~\ref{fig:comp_graph}.}{Figure~\ref{fig:comp_graph} illustrates the computational graph of this work.}
\section{Physics informed Neural Networks}\label{sec:pinns}Physics informed machine learning leverages \glspl{NN} to approximate solutions to partial differential equations~\cite{Raissi2019}.
In the \gls{PINN} framework,a stationary \gls{PDE} is formulated as
\begin{align}\begin{cases}\mathcal{N}u(\mathbf{x}) &=f(\mathbf{x}),\quad \mathbf{x} \in \Omega\\
\mathcal{B}u(\mathbf{x}) &=g(\mathbf{x}),\quad \mathbf{x} \in \partial\Omega
\end{cases}\label{eqn:PiNN_general}\end{align}on the (fully) connected domain $\Omega \in \mathbb{R}^d$ bounded by $\partial\Omega$.
The partial differential operator of the PDE is $\mathcal{N}$ and the \gls{PDE} can be of hyperbolic,elliptic,parabolic or mixed type.
$\mathcal{B}$ denotes a Dirichlet,Neumann or Robin boundary condition operator on $\partial\Omega$.
$f(\mathbf{x})$ is some homogeneous or inhomogeneous function and $g(\mathbf{x})$ are the boundary conditions.
The \gls{PDE} is solved on coordinate points in dimension $d$: $\mathbf{x}=\{x_1,x_2,\cdots,x_d\}^\mathsf{T} \in \Omega \in \mathbb{R}^d$ which are the argument of some function $u(\mathbf{x})=u_\nu(\mathbf{x})$ with parameters $\nu$.\newline
A quadratic residual function is then constructed as a target
\begin{align}\mathcal{L}_\mathrm{PINN,\mathcal{B}}=\mathop{\mathbb{E}}_{\mathbf{x} \in \Omega} &||\mathcal{N}u_\nu(\mathbf{x}) - f(\mathbf{x})||\nonumber\\
+ \mathop{\mathbb{E}}_{\mathbf{x}  \in \partial \Omega} &||\mathcal{B} u_\nu(\mathbf{x}) - g(\mathbf{x})||
\label{eqn:pinn_generic_loss}\end{align}where $\mathbb{E}(x)$ denotes the expectation operator.
Then,the optimization problem for the optimal parameters $\nu^{\star}$ becomes
\begin{equation}\nu^{\star}=\argmin_{\nu} \quad \mathcal{L}_\mathrm{PINN,\mathcal{B}}(\nu).
\label{eqn:pinn_optimization_general}\end{equation}Solving a \gls{PDE} as an optimization problem introduces challenges from optimization into the minimization of $\mathcal{L}_\mathrm{PINN,\mathcal{B}}$,for example,getting stuck at local minima.\newline
A common simplification to solve the optimization problem~\eqref{eqn:pinn_optimization_general} is the exact imposition of boundary conditions by modifying the output $u_\nu(\mathbf{x})$.
Constructing a monotonic function $\Phi(\mathbf{x})$ along the inward pointing normal vector on $\partial \Omega$ with approximate distance functions,such that $\Phi(\mathbf{x})=0 \,\,\forall \,\mathbf{x}\in\partial\Omega$ and $\Phi(\mathbf{x})>0 \,\,\forall \,\mathbf{x}\in\Omega$,decreases computational cost and can increase the final accuracy for many combinations of \glspl{PDE} and \glspl{PINN}~\cite{Sukumar2022}.
In the case of homogeneous Dirichlet boundary conditions,the second part of the loss function~\eqref{eqn:pinn_generic_loss} is removed by modifying
\begin{equation}\bar{u}_\nu(\mathbf{x})=g(\mathbf{x}) + \Phi(\mathbf{x})u_\nu(\mathbf{x})
\label{eqn:pinn_adf}\end{equation}to exactly satisfy the boundary condition for $\mathbf{x}\in\partial\Omega$.
In terms of equation~\eqref{eqn:pinn_generic_loss} this step reduces the loss function to $\mathcal{L}_\mathrm{PINN}=\mathop{\mathbb{E}}_{\mathbf{x} \in \Omega} ||\mathcal{N}u_\nu(\mathbf{x}) - f(\mathbf{x})||$ and removes issues associated with weighting boundary loss against domain loss. The presented approach will only use this type of loss formulation.\newline
The name \textit{Physics-informed Neural Network} specifically refers to one or more \glspl{NN} with parameters optimized to minimize some \gls{PDE} residual.
They do not require labelled data,such that the concept of \gls{PINN}'s may be viewed as an unsupervised learning strategy~\cite{Cuomo2022}.
\subsection{PINNs applied to the 3D ideal MHD equilibrium problem}\label{sec:pinn_mhs}\replaced[id=TT]{Our}{The} \gls{PINN} approach solves the stellarator-symmetric ideal \gls{MHD} equilibrium problem in the spectral representation~(\ref{eqn:R_fourier}-\ref{eqn:Z_fourier})\deleted[id=TT]{which also serves as the interface between \texttt{VMEC} and other codes}.
It replaces \texttt{VMEC}'s finite-difference with automatic differentiation gradients and moves the parameters to the weights and biases of three \glspl{NN},one for each dependent coordinate $NN_{X,mn}(f(\rho);\pmb{\nu}_X):\mathbb{R}\,\cap\,(0,1) \rightarrow \mathbb{R}^{M(2N+1)-N}$ with parameters $\pmb{\nu}_X$ for $X\in\{R,\lambda,Z\}$\deleted[id=TT]{ and two hidden layers}.
\added[id=TT]{The square root of \texttt{VMEC}'s radial coordinate,$\rho=\sqrt{s}=\sqrt{\frac{\psi}{\psi_\mathrm{b}}}$,which coincides with \texttt{DESC}'s radial coordinate,is used as \gls{NN} input.}
We briefly tested \glspl{NN} with only one hidden layer but our results were not performant compared to two hidden layers.
\replaced[id=TT]{Two-hidden-layer \gls{MLP} serve}{This serves} as the map from the radial coordinate $\mathbf{x} \coloneq \rho \,\in (0,1)$ to stellarator-symmetric Fourier modes $X_{mn}=\{R_{mn}^\mathrm{c},\lambda_{mn}^\mathrm{s},Z_{mn}^\mathrm{s}\}$.
The axis ($\rho=0$) and boundary ($\rho=1$) are not included during optimization.
$NN_{\mathrm{R},mn}$ and $NN_{\mathrm{Z},mn}$ represent the Fourier modes with a distance function~$\Phi(\rho)=(1-\rho^2)$ akin to equation~\eqref{eqn:pinn_adf}\begin{equation}X_{mn}(\rho)=X_{\mathrm{b}mn} + (1-\rho^2) NN_{X,mn}(f(\rho); \pmb{\nu}_X).
\end{equation}\noindent
This ensures that the modes coincide with the prescribed boundary modes at $\rho=1$,thereby reducing the dimensionality of the optimization space.
The boundary modes for $R$ and $Z$ are specified up to some maximum poloidal $M_\mathrm{b}$ and toroidal $N_\mathrm{b}$ mode number.
$\lambda_{mn}$ profiles are free of this restriction.
Interpolation of $m=0$ modes between axis and boundary serves as initial guess enforced via the last layer's bias of $NN_{\mathrm{R},0n}$ and $NN_{\mathrm{Z},0n}$:
\begin{equation}b_{L,X,0n}=X_{\mathrm{a}0n} - X_{0n}(\rho=0) \quad X \in \{R,Z\}\label{eqn:mhdinn_initial_guess}\end{equation}Initializing the Fourier modes for $R$ and $Z$ this way intends to roughly reproduce \texttt{VMEC}'s initialization by assuming mode-profiles of the form $X_{mn}=s X_{\mathrm{a}mn} + (1-s)X_{\mathrm{b}mn}$,ignoring some scaling factors used in \texttt{VMEC}.
Modes with $m\geq1$ require a fit at initialization because of the dependence on $\rho^m$ in equations~\eqref{eqn:R_NN_pred} and~\eqref{eqn:Z_NN_pred} and are set to $0$.
Subtracting the initial axis modes $NN_{X,0n}(\rho=0)$,this scheme can accommodate initial \gls{NN} weights,which are sampled from a normal distribution,with magnitudes of order $\mathcal{O}\sim10$ and higher without overlapping flux surfaces.\newline
If an initial guess constructed with the mimicked \texttt{VMEC} interpolation is not nested or initial axis modes are not provided,a new initial guess for the axis $X_{\mathrm{a}0n}$ is computed using an invertible mapping to boundary conforming coordinates~\cite{Babin2025}.
Given $\iota(\rho)$,$p(\rho)$ and $X_{\mathrm{b}mn}$,the solver then optimizes the weights of each \gls{NN} by minimizing
\begin{align}\mathcal{L}_\mathrm{PINN}=& \frac{1}{2} \sum_i (\,||\mathbf{F}(\rho)||_2^2 \,)_i.    \label{eqn:mhdinn_loss}\end{align}where $\mathbf{F}(\rho)$ is the ideal \gls{MHD} force residual~\eqref{eqn:f_magn} which is summed over all grid points \added[id=TT]{$i$}.\newline
The residual~\eqref{eqn:mhdinn_loss} is optimized by two optimizers: The initial minimization is performed by a first-order optimizer (\texttt{AdamW}),and the second minimization uses optimizers that approximate the Hessian (\texttt{BFGS}) and is stopped when the parameters do not change anymore or upon reaching some $\fvolnorm$ (only used in figure~\ref{fig:scan_until_DESC_fvol} for the \gls{NN} solutions).\newline
Training of \gls{NN} is often split into \textit{batches} but here we do not split the geometric grid and train on separate batches,but optimize on all grid points at once in each iteration step.
\newpage
\begin{strip}\noindent\rule{0.5\textwidth}{1pt}\begin{align}R(\rho,\theta,\zeta) &=\sum_{m=0,n=-N}^{M-1,N} \rho^m [R_{\mathrm{b}mn} + (1-\rho^2) NN_{\mathrm{R},mn} (f(\rho); \pmb{\nu}_R)] \,\cos(\varphi_{\mathrm{F},mn}) \label{eqn:R_NN_pred}\\[3pt]
\lambda(\rho,\theta,\zeta) &=\sum_{m=0,n=-N}^{M-1,N} \rho^m [NN_{\lambda,mn} (f(\rho); \pmb{\nu}{}_\lambda)] \,\sin(\varphi_{\mathrm{F},mn}) \label{eqn:L_NN_pred} \\[3pt]
Z(\rho,\theta,\zeta) &=\sum_{m=0,n=-N}^{M-1,N} \rho^m [Z_{\mathrm{b}mn} + (1-\rho^2) NN_{\mathrm{Z},mn} (f(\rho); \pmb{\nu}_Z)] \,\sin(\varphi_{\mathrm{F},mn}) \label{eqn:Z_NN_pred}\end{align}\noindent\hspace*{0.5\textwidth}\rule{0.5\textwidth}{1pt}\end{strip}Independent spectral resolutions in $M$ and $N$ for each dependent vector of modes $R_{mn}^\mathrm{c}$,$\lambda_{mn}^\mathrm{s}$ and $Z_{mn}^\mathrm{s}$ are configurable in the solver,but not investigated in this study.
A more detailed description of the \glspl{NN} is provided in~\ref{asec:NN}.\newline
Continuation methods that change the geometric or spectral resolutions as common practice in \texttt{DESC} (section~\ref{sec:desc}) or \texttt{VMEC} (section~\ref{sec:vmec}) are not used.
\section{Results}\label{sec:results}\subsection{Methodology}\label{sec:methodology}This section compares the numerical results of \texttt{VMEC} and \texttt{DESC} with the \gls{NN}-based results in terms of the force residual~\eqref{eqn:f_abs_norm}.
The goal of this comparison is to highlight the ability of the \glspl{NN} to be the parametrization to compute some minimum force error on equal spectral resolution for iota-prescribed,fixed-boundary equilibria.
A more detailed runtime comparison between \texttt{DESC} and the \gls{NN}-based results concludes this section.\newline
The residual is normalized with the volume averaged magnitude of the magnetic pressure gradient~\eqref{eqn:b_magnitude} \deleted[id=TT]{to ensure a fair comparison between equilibria}\begin{equation}\mathbf{F}_\mathrm{norm}=\frac{|\mathbf{J} \times \mathbf{B} - \nabla p|}{\,\,\langle |\nabla |B|^{2}/(2\mu_0)| \rangle_\mathrm{vol}}
\label{eqn:f_abs_norm}\end{equation}\added[id=TT]{which is constant for each shown equilibrium,but we still normalize the force error with it to facilitate comparisons between different equilibria.}All shown equilibria are prescribed with power profiles for $\iota(\rho)$ with coefficients $a_l$ and $p(\rho)$ with coefficients $a_m$: $p(\rho)=\sum_m \rho^{2m} a_m$ and analogous for $\iota(\rho)$.\newline
\texttt{VMEC} serves as a qualitative benchmark for magnetic flux surface topology.
In the following,we compare the flux surface geometry of the \gls{NN}-based solution to those of \texttt{VMEC} by visually overlaying both Poincaré plots at $\zeta=0$ in figure~\ref{fig:DshapePoincare} and~\ref{fig:W7XPoincare} for ten flux surfaces including the last surface.
Each Poincaré comparison also shows arcs of the straight field line poloidal angle $\theta^{\star}=\theta + \lambda(\rho,\theta,\zeta)$ at eight equidistant locations.\newline
The normalized force residual~\eqref{eqn:f_abs_norm} offers a more reliable,quantitative metric,and its flux surface average $\langle \mathbf{F}_\mathrm{norm} \rangle$ is plotted next to each Poincaré comparison for the three solvers.
Computational runtime and accuracy in terms of the volume averaged $\langle \mathbf{F}_\mathrm{norm} \rangle$ are compared only between \texttt{DESC} and \glspl{NN},because \texttt{DESC} and the \gls{NN}-based approach need to be run on modern graphical accelerators to compute high resolution 3D equilibria in reasonable time and because both solvers compute solutions without \texttt{VMEC}'s force residual spike.\newline
Volume averaging $\mathbf{F}_\mathrm{norm}$ provides a scalar value to compare different solutions of one equilibrium
\begin{align}\fvolnorm &=\langle \mathbf{F}_\mathrm{norm} \rangle_\mathrm{vol}\nonumber \\
&=\frac{1}{V} \int \mathbf{F}_\mathrm{norm} \sqrt{g} d\pmb{\alpha}.
\label{eqn:f_vol_norm}\end{align}The equilibrium solutions shown in this work are calculated using \texttt{VMEC}'s or \texttt{DESC}'s convention for the smallest poloidal and toroidal resolution of the computational grid given some spectral resolution.
The \gls{NN}-based approach uses $50$ flux surfaces for figure~\ref{fig:DshapePoincare} and $32$ for figure~\ref{fig:W7XPoincare} without refinement and without exploiting stellarator up-down symmetry,that is $\theta \in [0,2\pi)$ for the \gls{NN}'s computational grid,\added[id=TT]{but in terms of Fourier modes,stellarator symmetry is exploited (see section~\ref{subsec:3d-magnetohydrodynamic-equilibrium})}.
All collocation points are equidistant in $(\rho,\theta,\zeta)$ and are not changed during optimization for the \glspl{NN}\added[id=TT]{ because the \texttt{DESC} solutions,to which we compare the \gls{NN} approach,use a constant grid as well.}
For all equilibria we use the minimal amount of grid points recommended to run \texttt{VMEC},which is $2M+6$ poloidal grid points and $2N+4$ toroidal grid points for the \gls{NN} solutions.
Exact parameters can be found in the supplemental material and a mathematical description of the \gls{MLP} is provided in~\ref{asec:NN}.
The computational grids in \texttt{DESC} use double the amount of grid points for each $L_{ZP}$,$M_D$ and $N$.\newline
The minimization of each \texttt{DESC} solution follows a two-step procedure using the least-squares optimizer:
First,the solution is computed using the default continuation algorithm with default tolerances.
A secondary solve (at $\eta_\mathrm{b}=1$ and $\eta_\mathrm{p}=1$) then reduces the optimizer tolerances to $0$,approximating machine precision.
Default geometric and spectral resolutions are used in \texttt{DESC},and the axisymmetric tokamak is solved with fringe spectral indexing of the Zernike polynomial while \gls{W7X} is solved with ansi indexing.\newline
Examples for fixed-boundary equilibria are,on the one hand,a tokamak taken from~\textcite{Hirshman},and,on the other hand,a $M_\mathrm{b}=N_\mathrm{b}=12$ \gls{W7X} equilibrium in standard configuration presented in~\textcite{Panici2023}.
Both equilibria come with an initial guess for the magnetic axis.
Many other fixed-boundary,finite-$\beta$ equilibria can be compared,but we choose these two equilibrium cases because they have already been compared between \texttt{VMEC} and \texttt{DESC}~\cite{Panici2023}.\newline
The remainder of the results compare \texttt{DESC} and various sizes of the two-hidden-layer \glspl{NN},using the \gls{W7X} equilibrium with boundary harmonics truncated to $M_\mathrm{b}=N_\mathrm{b}=10$.
Figure~\ref{fig:scan_time_w7xM10N10} shows $\fvolnorm$ compared to \texttt{DESC} for the truncated \gls{W7X} equilibrium plotted over wall-time (both codes used a Nvidia A100 Graphical Processing Unit (GPU)).
The same datapoints of figure~\ref{fig:scan_time_w7xM10N10} are used in figure~\ref{fig:scan_mn_w7xM10N10} but with the global spectral resolution $M=N$ of the solvers on the x-axis.
Finally,figure~\ref{fig:scan_until_DESC_fvol} shows the training of \glspl{NN} with layer widths $n_\mathrm{l}=\{1,2,4,8,16\}$ until the \glspl{NN} $\fvolnorm$ matches the minimum $\fvolnorm$ computable with \texttt{DESC} for the boundary-truncated \gls{W7X} equilibrium.
The \gls{NN} minimization uses only $32$ flux surfaces in this comparison.
Results for a Heliotron-like equilibrium as used in~\textcite{Hirshman} and~\textcite{Dudt2020} are provided in the supplementary material,because analysis of possibly bifurcated equilibrium states would take considerably more effort.\newline
We provide these comparisons not to create a notion of a \textit{best} solver,because this is essentially not possible respecting all differences of the three solvers,but to show that the \gls{NN}-based approach is competitive in time and minimum computable residual to modern solvers.\newline
All hyperparameters for optimizers and \gls{NN} and result data for each solver and equilibrium are provided in the supplementary material (see~\ref{asec:data}).
\subsection{Equilibrium solutions}
\begin{figure*}
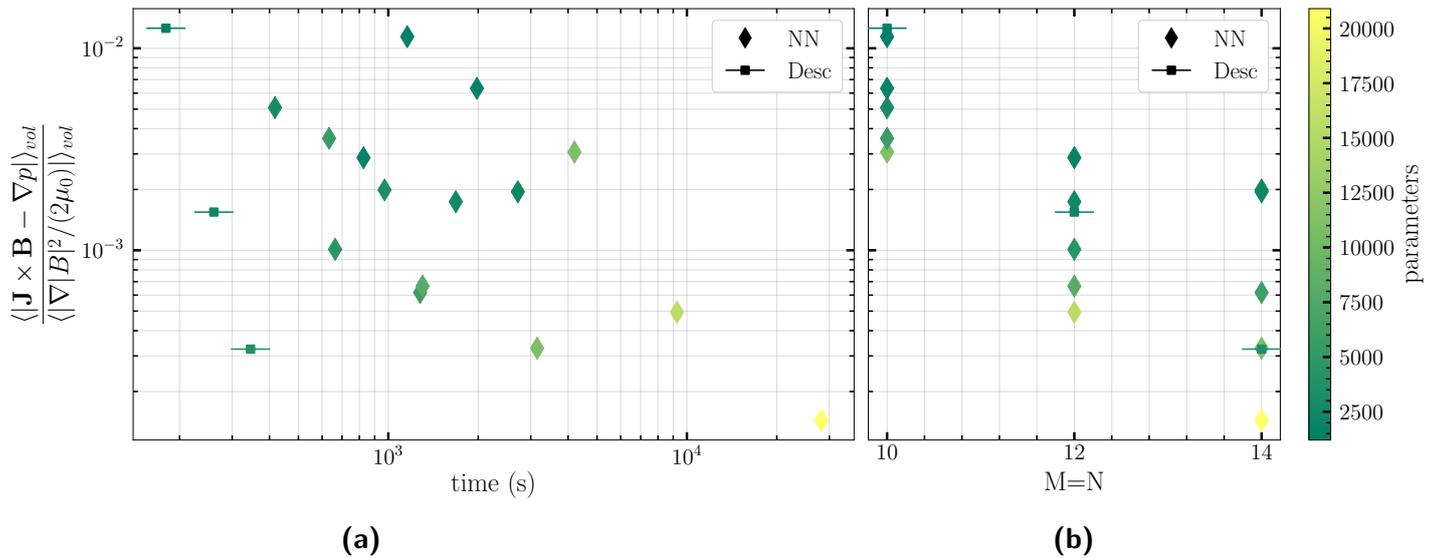
\captionsetup{width=1\textwidth}\centering
\igraph[width=1\linewidth]{content_figures_Both_log_regress_Fvol_MN.pgf}\begin{subfigure}[b]{0.5\textwidth}\caption{ }\label{fig:scan_time_w7xM10N10}\end{subfigure}\hfill
\begin{subfigure}[b]{0.5\textwidth}\caption{ }\label{fig:scan_mn_w7xM10N10}\end{subfigure}\caption{Minimum force error for \gls{W7X} standard configuration with $M_\mathrm{b}=N_\mathrm{b}=10$ boundary harmonics with global resolutions $M=N=\{10,12,14\}$ computable by \texttt{DESC} and \glspl{NN} with layer widths of $n_\mathrm{l}=\{1,2,4,8,16\}$. The colorbar indicates the number of parameters and the y-axis the normalized volume averaged force error~\eqref{eqn:f_vol_norm} plotted over (a) time in seconds and (b) Fourier resolution parameter $M=N$.}
\label{fig:both_regress_mn}\end{figure*}

Figure~\ref{fig:DshapePoincare} displays a Poincaré comparison between \glspl{NN} and \texttt{VMEC} of an axisymmetric,D-shaped tokamak with $\beta \sim 3\%$,computed with $M=11$ poloidal modes.
\texttt{VMEC}'s grid is refined up to $2048$ flux surfaces and reaches a final,minimum internal tolerance of $2\times10^{-16}$.
We use this number of flux surfaces in \texttt{VMEC} because iterative refinement of the radial grid is a common technique to use in \texttt{VMEC} minimizations and this comparison focuses on equal spectral resolution.
Each of the three \glspl{NN} for this equilibrium has two hidden layers with $n_\mathrm{l}=8$ nodes.
The boundary geometry of the D-shaped tokamak is given by
\begin{align}R_\mathrm{b} &=3.51 - \cos ( \theta ) + 0.106 \,\cos ( 2 \theta )\nonumber \\
Z_\mathrm{b} &=1.47 \,\sin (\theta) + 0.16 \,\sin ( 2 \theta)
\label{eqn:dshape_def}\end{align}with input pressure profile $p(\rho)=1600 \,( 1 - \rho^2 )^2$,rotational transform $\iota(\rho)=1 - 0.67 \,\rho^2$ and total toroidal flux set to $\psi_\mathrm{b}=1$ \si{Wb}~\cite{Hirshman}.
The flux surfaces of \texttt{VMEC} are plotted in red and the flux surfaces of the \glspl{NN} in dashed green.
\textcite{Dudt2020} found a similarly good agreement between the flux surfaces as computed by \texttt{VMEC} and \texttt{DESC}.
All three solvers converge to a similar location for the magnetic axis and qualitatively equal $\theta^{\star}$.
The \glspl{NN} improves the flux surface averaged force error compared to \texttt{DESC} at this spectral resolution (figure~\ref{fig:dshapeFfsa}),and both codes approach the same value close to axis.
The force residual of \texttt{VMEC} displays the largest discrepancies in the vicinity of the axis and close to the boundary compared to the \glspl{NN} and \texttt{DESC}.
The normalized residual of \texttt{VMEC} rises to $\sim 10^{-1}$ at $\rho<0.15$ while \texttt{DESC} and the \gls{NN}-based solution do not show this spike close to the axis (see~\ref{asec:analytic_constr}).\newline
To compare the three solvers for a non-axisymmetric case,we choose a \gls{W7X} equilibrium,akin to the standard configuration,with $M_\mathrm{b}=N_\mathrm{b}=12$ boundary harmonics.
\textcite{Panici2023} compares \texttt{DESC} and \texttt{VMEC} solutions for this equilibrium,and we employ the same input specification used in that work for this analysis.
\texttt{VMEC}'s radial grid for this equilibrium is refined up to $1024$ flux surfaces and the \glspl{NN} use $n_\mathrm{l}=18$.
The same trend as in the D-shaped tokamak solutions is visible in figures~\ref{fig:W7XPoincare} and~\ref{fig:W7XFfsa}.
Flux surfaces,$\theta^{\star}$ and axis position agree qualitatively for \texttt{VMEC} and \glspl{NN},and their difference becomes most visible for the force residual $\langle \mathbf{F}_\mathrm{norm} \rangle$ close to axis.
\texttt{VMEC}`s force residual increases at $\rho<0.15$ while \texttt{DESC} and \glspl{NN} converge to solutions without this spike.
The \gls{NN}'s $\langle \mathbf{F}_\mathrm{norm} \rangle$ decreases towards the axis compared to \texttt{DESC}\@.\newline
Of interest to users of \gls{MHD} equilibrium solvers is not only their physical validity but also their speed to find accurate solutions.
Figure~\ref{fig:scan_time_w7xM10N10} to~\ref{fig:scan_until_DESC_fvol} compare \texttt{DESC} solutions with \gls{NN} solutions for multiple cases of the truncated \gls{W7X} equilibrium.
To perform scans on memory-limited hardware,the boundary of the equilibrium used in figures~\ref{fig:scan_time_w7xM10N10} to~\ref{fig:scan_until_DESC_fvol} is pruned to only include modes up to $M_\mathrm{b}=N_\mathrm{b}=10$.
Figure~\ref{fig:scan_time_w7xM10N10} and figure~\ref{fig:scan_mn_w7xM10N10} compare the same data plotted against time on plot~\ref{fig:scan_time_w7xM10N10} and against internal resolution parameter $M=N=\{10,12,14\}$ on plot~\ref{fig:scan_mn_w7xM10N10} to the right.
The datapoints each describe one instantiation of the solver run until the optimizer does not update the parameters anymore; i.e.~the optimizer is fully converged.
Individual \texttt{DESC} runs are marked with a crossed square and \gls{NN} results with a rhombus.
\gls{NN}-based solutions are plotted with layer widths of $n_\mathrm{l}=\{1,2,4,8,16\}$ for each of $M=N=\{10,12,14\}$.
\begin{figure*}[!htb]
\begin{minipage}[t]{0.5\linewidth}\centering
\scalebox{0.8}{\input{content_figures_dshapePoincareNoF.pgf}}
\caption{Poincare sections for a D-shaped tokamak test-case with boundary defined by~\eqref{eqn:dshape_def}. Dashed green lines represent the \glspl{NN} solution with $n_\mathrm{l}=8$ hidden nodes and full red lines represent the \texttt{VMEC} solution,revealing a qualitative agreement in flux surface geometry. Profiles of $\theta^{\star}=\theta + \lambda(\rho,\theta,\zeta)$ are plotted at eight $\theta^{\star}$ locations from axis to boundary.
}\label{fig:DshapePoincare}\end{minipage}\begin{minipage}[t]{0.5\linewidth}\centering
\scalebox{0.65}{\input{content_figures_dshape_ns50_ffsa.pgf}}
\caption{Normalized flux surface averaged force error for the solutions of \texttt{VMEC},\glspl{NN} and the \texttt{DESC} for the D-shaped tokamak test-case plotted over $\rho$. The \gls{NN} optimization continued for about 100 times longer than \texttt{DESC} and \texttt{VMEC} but achieves lowest minimum force error over the whole volume. \texttt{VMEC}'s force error spikes at $\rho<0.2$.}\label{fig:dshapeFfsa}\end{minipage}\end{figure*}\begin{figure*}[!htb]
\vspace{2em}\begin{minipage}[t]{0.5\linewidth}\centering
\scalebox{0.8}{\input{content_figures_w7xM11N12Poincare.pgf}}
\caption{Poincare section for a $M_\mathrm{b}=N_\mathrm{b}=12$ equilibrium of the \gls{W7X} stellarator solved with \glspl{NN} with $n_\mathrm{l}=18$ hidden nodes in dashed green and \texttt{VMEC} in red. Both solutions agree with each other qualitatively. $\theta^{\star}=\theta + \lambda(\rho,\theta,\zeta)$ profiles are plotted at eight $\theta^{\star}$ locations from axis to boundary.}\label{fig:W7XPoincare}\end{minipage}\begin{minipage}[t]{0.5\linewidth}\centering
\scalebox{0.65}{\input{content_figures_w7xM11N12_F_fsa.pgf}}
\caption{Normalized flux surface averaged force error for \texttt{VMEC},\glspl{NN} and \texttt{DESC} for the $M_\mathrm{b}=N_\mathrm{b}=12$ \gls{W7X} test-case plotted over $\rho$. The \glspl{NN} achieve lowest minimum force error over the whole volume and the force residual of \texttt{VMEC} spikes at $\rho<0.15$.}\label{fig:W7XFfsa}\end{minipage}\end{figure*}\newpage
\mbox{}\newpage
\mbox{}\newpage
On the same hardware,\texttt{DESC} computes its solutions faster,but with a higher final force residual compared to some of the layer width of the two-hidden-layer \glspl{NN}\@.
The \glspl{NN} are able to achieve solutions with lower $\fvolnorm$ but at increased computational cost,which can be easily reduced by,for example,using a grid with $\theta\in[0,\pi)$ that removes duplicated nodes under stellarator symmetry.
With increasing spectral resolution,both solvers show an exponential decreasing $\fvolnorm$ in figure~\ref{fig:scan_mn_w7xM10N10}.\newline
To answer the question of how \texttt{DESC} and the \glspl{NN} compare in terms of compute resources,the $M_\mathrm{b}=N_\mathrm{b}=10$ \gls{W7X} case is solved using \texttt{DESC} with the two-stage approach described in section~\ref{sec:methodology} until parameters stop changing.
The minimum $\fvolnorm$ computable by \texttt{DESC} is then set as a stopping criterion for the \gls{NN} training process,which computes equilibria that are within $0.5\%$ relative tolerance with respect to this stopping criterion.
The results of this test are plotted in figure~\ref{fig:scan_until_DESC_fvol} over the compute time in seconds on the x-axis.
\gls{NN}-based solutions achieve the same force error requiring more computational time,node number of the hidden layers{layer width}.
Notably,the \gls{NN} training procedure has an additional slight overhead to compute $\fvolnorm$ at each optimization step to check if the stopping condition is fulfilled.
\begin{figure}\centering
\igraph[width=0.9\linewidth]{content_figures_w7x_mn10_speed.pgf}\caption{\texttt{DESC} trained until convergence for the \gls{W7X} case pruned to $M_\mathrm{b}=N_\mathrm{b}=10$ boundary harmonics. \gls{NN} with layer widths $n_\mathrm{l}=\{2,4,8,16\}$ trained until \texttt{DESC}'s $\fvolnorm$ reached within relative tolerance of $0.5\%$. Both codes are executed on the same GPU.}
\label{fig:scan_until_DESC_fvol}\end{figure}\section{Discussion}\label{sec:discussion}The \gls{NN} parametrization can be optimized to represent equilibrium solutions with lower residual compared to \texttt{VMEC} and \texttt{DESC} for most node numbers of the two hidden layers for the two presented equilibrium cases and the Heliotron equilibrium in the supplemental material (see~\ref{asec:data}).
Both the Fourier Zernike basis of \texttt{DESC} and the \gls{NN} parametrization satisfy the analytic constraint on functions that represent physical scalars on the unit disc~\eqref{eqn:analyticity},improving upon \texttt{VMEC} by removing the force error spike towards the axis (see figures~\ref{fig:dshapeFfsa},~\ref{fig:W7XFfsa} and~\ref{asec:analytic_constr}).\newline
Optimizing the \gls{NN} until the minimum computable force error of \texttt{DESC} incurs comparable computational cost (figure~\ref{fig:scan_until_DESC_fvol}).
But with more resources,the \gls{NN}-based approach computes lowest force residuals for the tested equilibria and solvers (figures~\ref{fig:W7XFfsa},~\ref{fig:dshapeFfsa} and~\ref{fig:scan_mn_w7xM10N10}).
It is unexpected that the \gls{NN} approach is comparable in terms of cost given the tendency of \glspl{NN} to get stuck in local minima,the naive initial weights drawn from some normal distribution and the common optimizers used in this work.
Remarkably,the presented \gls{NN} approach does not use continuation methods as used in \texttt{DESC} and neither does it use spectrally condensed representations as used in \texttt{VMEC}.
The presented differences in computational cost could stem from using \glspl{NN} or could be attributable to implementation nuances; this question can be answered by comparing only floating point operations to reach a certain residual of each code.\newline
Training procedures of \glspl{PINN} usually fit the low frequencies of the solution first,also known as spectral bias~\cite{Kiessling2022},and the presented approach could benefit from it in terms of equation~\eqref{eqn:spectral_width}.
One possible reason for the lower minimum force error is the interplay of optimizer with the truncated representation used in \texttt{VMEC} and \texttt{DESC}.
Representing solutions to ideal MHD with truncated spectral resolution regularizes the equilibrium problem~\cite{Burby2023},which is why the results of this work might be explained by the universal function approximation theorem for deep narrow \glspl{NN}~\cite{Kidger2020}.
The distance between $\fvolnorm$ of \texttt{DESC} and the \glspl{NN} decreases with increasing spectral resolution (see figure~\ref{fig:scan_mn_w7xM10N10}),adding another support to this explanation.
As this efficiency comparison only relies on one equilibrium of \gls{W7X} in standard-configuration,this effect should be studied for many different equilibria and \gls{NN} models.\newline
Section~\ref{sec:results} compares \texttt{VMEC},\texttt{DESC} and the \gls{NN}-based approach on small spectral and geometric resolutions to show the viability of the \gls{NN}-based parametrization.
Increasing the grid resolution of \texttt{DESC} could result in lower force residuals and running \texttt{VMEC} multiple times,using the axis of the previous run as initial guess for the next run,could reduce \texttt{VMEC}'s force residual further.
This work focuses on performance and solution quality given low resolutions and single solves; therefore,improvements to \texttt{DESC}'s geometric or spectral resolution and \texttt{VMEC}'s resolutions or workflows are not explored.
The number of parameters in \texttt{DESC} is tightly coupled to the chosen spectral parameters $L_\mathrm{ZP}$,$M_\mathrm{D}$ and $N$.
In \texttt{VMEC},the number of parameters changes depending on the number of flux surfaces,$M$ and $N$.
The \gls{NN}-based approach is the only one of the three solvers for which the number of parameters can be set independently of the resolution of the spectral basis or the geometric grid size.
Understanding the advantages and disadvantages of this independence can lead to even faster and more accurate solvers and more sophisticated pre-conditioning methods applicable to all ideal \gls{MHD} solvers.\newline
The presented work provides value for future operator learning studies and the reduced residual for higher node numbers in the two hidden layers $n_\mathrm{l}$ shows benefits of using small \glspl{NN} to parametrize Fourier modes of ideal \gls{MHD} equilibria.
\subsection{Future Work}Halving the grid in the poloidal coordinate,i.e.~$\theta \in [0,\pi)$,removes collocation points which encode equal information in stellarator symmetry and is expected to be an enhancement that likely will increase the efficiency of the presented approach further.\newline
\Gls{MHD} solvers are extensively used as inner loop in stellarator optimization and the normalized force residual is a metric for ideal \gls{MHD} equilibrium solutions.
Equilibria resulting from the outer stellarator optimization loop are implicitly determined by the individual solutions of the inner loop.
Many solvers that compute stability,transport or turbulence metrics use the resulting equilibrium magnetic field of the outer loop,motivating a sensitivity analysis of those metrics with respect to the residual of the inner loop's \gls{MHD} equilibrium solution.
Computing these metrics for a single equilibrium solved by multiple solvers can help to validate the \gls{NN}-based approach and provide an initial guess for their sensitivity with respect to the force error and solver \added[id=TT]{numerics}.
Costs associated with solving the ideal \gls{MHD} equilibrium are tiny compared to stability,transport or turbulence calculations and those dwarf compared to the cost of building large devices,but lowering the force residual of the inner loop could,however,improve Pareto fronts in multi-objective stellarator optimization.\newline
Quantifying information overlap between \gls{NN} parametrizations of different but related equilibria can unlock faster and more precise convergence or pre-conditioning schemes and is a prerequisite for transfer learning and helpful for the design of operator models.
Analyzing the basis function encoded by the \glspl{NN} \deleted[id=TT]{in this work} is \added[id=TT]{not only} a first step in that direction \added[id=TT]{but has the potential to improve ideal MHD solvers in general}.\newline
Integration of these novel \gls{NN}-based equilibrium solutions into the framework of the \texttt{DESC} code provides a test bench in terms of many stellarator optimization metrics implemented in \texttt{DESC}.
This integration would shed some light on the question whether the lower residuals presented in this work are attributable to using \glspl{NN} and Fourier basis,or whether the Fourier-Zernike basis of \texttt{DESC} with small spectral resolution upper-bounds force residuals.\newline
Training a model for some operator on solution data in a continuous space of equilibria induces some epistemic error.
Replacing the \texttt{VMEC} dataset with a dataset created with the presented approach could reduce the epistemic error of the operator model introduced by~\textcite{Merlo2023} to acceptable levels for rapid stability analysis or other applications,for which the current models lacks precision.\newline
This work only varied the node number of two hidden layers (see figures~\ref{fig:scan_time_w7xM10N10},~\ref{fig:scan_mn_w7xM10N10} and~\ref{fig:scan_until_DESC_fvol}),but extending the search to varying the number of hidden layers and nodes could yield lower residuals.\newline
The \glspl{NN} hidden layer node numbers $n_\mathrm{l}$ are purposely kept minimal and \glspl{NN} with more capacity,on average,reach lower residuals for larger computational cost (see figure~\ref{fig:scan_mn_w7xM10N10}),hinting at potential to improve the approach both in resource requirements and time by leveraging \gls{NN} optimization techniques.
This includes,but is not limited to,a search over \gls{NN} structures commonly used in the domain of \glspl{PINN}~\cite{Wang2024},novel collocation point sampling techniques to increase optimization efficiency,an investigation into novel self-adaptive loss functions used in other \glspl{PINN},and more~\cite{Luo2025}.
\added[id=TT]{However,many of these enhancements to the current approach increases the likelihood of the minimization stagnating in local optima.}\newline
Investigating the interplay between spectral bias~\cite{Kiessling2022} and spectral width (see equation~\eqref{eqn:spectral_width}) could reveal if the \gls{NN}-based parametrization of ideal \gls{MHD} equilibria rather benefits from it instead of being handicapped.
Also,future work could explore whether the strong form of the ideal \gls{MHD} residual~\eqref{eqn:f_magn} is sufficient to remove spectral enhancements during optimization.\newline
A comparison between the codes based on floating point operations is out of scope for this work,but would shed light on the exact efficiencies of the three codes.
However,with ever decreasing computational costs per unit compute,the time of researchers might be better utilized to develop novel numerical methods.\newline
Another worthwhile investigation is whether the presented results can be achieved without any \glspl{NN},solely by optimizing the second order force operator~\eqref{eqn:f_vol_norm} over the parameters of the \texttt{VMEC}-like Fourier basis directly,using finite-difference gradients in radial direction.
However,this is only interesting for solving single equilibria and any operator model requires an additional interpolating model,e.g. the small \gls{MLP} used in this work.
\section{Conclusion }\label{sec:final}This work presents a novel approach to solve fixed-boundary,finite-$\beta$ ideal \gls{MHD} equilibria with nested flux surfaces and isotropic pressure using \glspl{NN}.
It is the first \gls{MHD} solver capable of adjusting its parametrization independently of the geometric or spectral resolution,and the results show some benefit to using higher hidden-layer node numbers $n_\mathrm{l}$.
\replaced[id=TT]{Residuals of \texttt{DESC} solutions and \gls{NN} solutions become increasingly similar with increasing spectral resolution. All \gls{NN} outperform \texttt{DESC} for $M=N=10$ but only the \gls{NN} with $n_\mathrm{l}=16$ outperforms \texttt{DESC} for $M=N=14$ in terms of accuracy and requires $\sim 2$ orders of magnitude more compute resources for a decrease in $\fvolnorm$ by a factor of $\sim 2$. For the equilibria and spectral resolutions we tested,at least one \gls{NN} parametrization resulted in the most accurate equilibrium solution. }{For all tested equilibria,this approach achieves the lowest force residual compared to other common ideal \gls{MHD} equilibrium solvers for most tested node numbers of the two-hidden-layer \glspl{NN}.}
Minimization of the residual parametrized by the \glspl{NN} until the minimum computable residual of the solver \texttt{DESC} for a \gls{W7X} equilibrium requires slightly more resources,but can,with higher computational cost,arrive at solutions with lowest residual for the tested solvers.
The deliberate choice of small \glspl{NN} in this \replaced[id=TT]{work}{initial exploration} establishes a lower bound for the complexity of \glspl{NN} representing a single equilibrium state \added[id=TT]{accurately},which could serve future work aimed at creating \gls{NN} models for \gls{MHD} operators.
The current optimization procedure is purposely kept minimal,but the literature surrounding \glspl{PINN} offers many improvements such as novel optimizers,new \gls{NN} structures and sampling techniques.
These enhancements are very likely to yield even lower residuals at lower cost.
The presented proof-of-principle of optimizing simple \glspl{NN} to minimize ideal \gls{MHD} residuals of a single 3D equilibrium provides a solid basis to extend these models to represent continuous,parametrized spaces of equilibria.
Possible applications include the optimization of devices for a whole continuum of configurations,acceleration of intra-shot analysis workflows that rely on the equilibrium magnetic topology and sophisticated control algorithms.
\ifthenelseproperty{compilation}{glossaries}{\ifthenelseproperty{compilation}{acronyms}{\printglossary[type=\acronymtype,style=mcoltree] }{}\ifthenelseproperty{compilation}{los}{\setlength\extrarowheight{5pt}\printglossary
[
title=List of Symbols,
type=symbols,
style=customListOfSymbols,
]
\setlength\extrarowheight{0pt}}{}}{}\ifthenelseproperty{compilation}{lof}{\input{templates_Default_lof.tex}}{}\ifthenelseproperty{compilation}{lot}{\input{templates_Default_lot.tex}}{}\ifthenelseproperty{compilation}{lol}{\input{templates_Default_lol.tex}}{}\ifthenelseproperty{compilation}{appendix}{\appendix
\renewcommand{\thesubsection}{Appendix \Alph{subsection}}
\section*{Appendix}\subsection{Data availability}\label{asec:data}Equilibria,models,some implementation files used in the presented work are available at~\hyperref[https://doi.org/10.5281/zenodo.15838028]{https://doi.org/10.5281/zenodo.15838028}.
The supplementary material also includes files that encode the presented \gls{NN} solutions in \texttt{VMEC}'s output file format. We also added plots showing the parallel current profiles and other plots of \texttt{VMEC},\texttt{DESC} and the \glspl{NN} to this material.\vspace{0.5em}\subsection{Neural Networks}\label{asec:NN}We intentionally use simple \glspl{NN} to provide a lower bound on \gls{NN} complexity:
The \glspl{NN} are two-layer ($L=2$) \glspl{MLP} with $\sigma=\tanh=(e^{2x}-1)/(e^{2x}+1)$ activation function and input $f(\rho)=2\rho^2-1 \in \mathbb{R}\,\cap\,(-1,1)$.
Each \gls{NN} is denoted by $NN_{X,mn}\,\,\text{for}\,\,X\in\{R,\lambda,Z\}$ and defined by a composition of affine transformations with nonlinearities
\begin{align}NN_{X,mn}(f(\rho)) &=W_{2,X}(\sigma(z_1(f(\rho))) + b_{2,X}\nonumber\\
z_1(f(\rho)) &=W_{1,X}(\sigma(z_0(f(\rho))) + b_{1,X}\nonumber\\
z_0(f(\rho)) &=W_{0,X}f(\rho) + b_{0,X}\nonumber
\end{align}with dimensions $W_{0,X}\in\mathbb{R}^{n_\mathrm{l}\times 1}$,$W_{1,X}\in\mathbb{R}^{n_\mathrm{l}\times n_\mathrm{l}}$,$W_{2,X}\in\mathbb{R}^{(M(2N+1)-N)\times n_\mathrm{l}}$,respective parameters $\pmb{\nu}_X$ and node number of the two hidden layers $n_\mathrm{l}$.
The summand of layers $l=\{0,1\}$ has dimension $b_{l,X}\in\mathbb{R}^{n_\mathrm{l}}$ and the last bias is of dimension $b_{2,X}\in\mathbb{R}^{M(2N+1)-N}$.
Parameters of the \gls{NN} weights $W_{l,X}$ are initialized by random sampling from a normal distribution with standard deviation $\sigma=0.01$ and mean $\mu=0$.
The bias $b_{l<L,X}$ are initially set to $0$ and $b_{L,X}$ are set such that the initial guess for the magnetic axis is satisfied (see section~\ref{sec:pinn_mhs}).\vspace{0.5em}\subsection{Analytic constraint}\label{asec:analytic_constr}The constraint on functions representing physical scalars with poloidal fourier modes $a_m$ near the axis of the unit disk~\cite{Lewis1990},namely that
$R_{mn}/\rho^m$ and $Z_{mn}/\rho^m$ do not blow up towards the axis,is
\begin{equation}\label{eqn:analyticity}a_m(\rho)=\rho^m \sum_i a_{m,i} \rho^{i} \,\,\,\text{for} \,i \in \{0,2,4,...\}.
\end{equation}For the axisymmetric solution shown in figures~\ref{fig:DshapePoincare} and~\ref{fig:dshapeFfsa},the profiles of $R_{mn}/\rho^m$ are plotted in figure~\ref{fig:dshape_analyticity}.
All equilibria solved by \glspl{NN} show no blow-up close to the axis for $R_{mn}/\rho^m$ and $Z_{mn}/\rho^m$.
\begin{figure}[H]
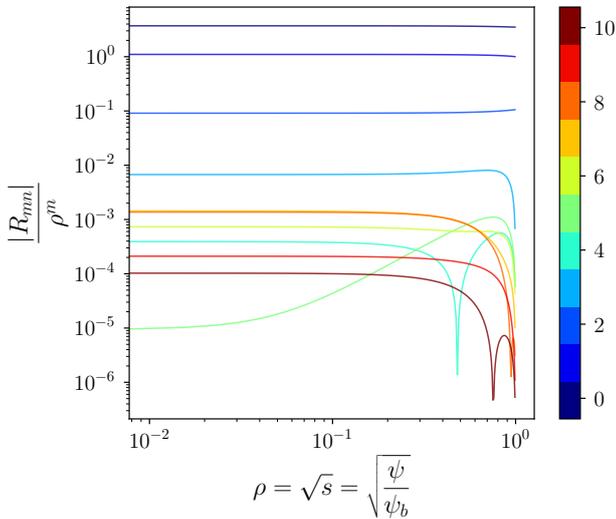

\centering
\igraph[width=0.9\linewidth]{content_figures_dshapeAnalyticity.pgf}\caption{$R_{mn}/\rho^m$ modes of \gls{NN} solution for axisymmetric D-shaped equilibrium.}
\label{fig:dshape_analyticity}\end{figure}\subsection{Training loss}\label{asec:train_loss}The evolution of $\mathcal{L}_\mathrm{PINN}$ (equation ~\eqref{eqn:mhdinn_loss}) for the \gls{W7X} equilibrium with $M=N=12$ poloidal and toroidal modes is illustrated in figure~\ref{fig:train_loss}.
\texttt{ADAM-W} was configured for $250000$ steps and \texttt{BFGS} was configured to stop after the relative change in $\mathcal{L}_\mathrm{PINN}$ after the past $10$ steps is below $10^{-10}$.
Despite requiring fewer steps,the \texttt{BFGS} optimizer required $1.5$ orders of magnitude more compute resources but also further reduced the loss by $\sim 2$ orders of magnitude.
\begin{figure}[H]
\centering
\igraph[width=\linewidth]{content_figures_loss_over_all_steps_w7x.pgf}\caption{Evolution of $\mathcal{L}_\mathrm{PINN}$ for the \gls{NN} over training steps in the \gls{W7X} equilibrium case shown in figures~\ref{fig:W7XPoincare} and~\ref{fig:W7XFfsa}. Optimizers are marked with arrows below the x-axis.}
\label{fig:train_loss}\end{figure}}{}\ifthenelseproperty{compilation}{listofpublications}{\begin{refsection}\newrefcontext[sorting=ndymdt]
\nocite{*}\ifthenelseproperty{compilation}{clsdefineschapter}{\ifKOMA \addchap[Publications as first author]{Publications as first author}\label{sec:publications_as_first_author}\else
\chapter[Publications as first author]{Publications as first author}\label{sec:publications_as_first_author}\fi
}{\ifKOMA \addsec[Publications as first author]{Publications as first author}\label{sec:publications_as_first_author}\else
\section[Publications as first author]{Publications as first author}\label{sec:publications_as_first_author}\fi
}
\printbibliography[
keyword=firstAuthor,
keyword=refereed,
heading=subbibliography,
title={Peer-reviewed articles},
resetnumbers=true
]
\end{refsection}\begin{refsection}\newrefcontext[sorting=ndymdt]
\nocite{*}\ifthenelseproperty{compilation}{clsdefineschapter}{\ifKOMA \addchap[Publications as coauthor]{Publications as coauthor}\label{sec:publications_as_coauthor}\else
\chapter[Publications as coauthor]{Publications as coauthor}\label{sec:publications_as_coauthor}\fi
}{\ifKOMA \addsec[Publications as coauthor]{Publications as coauthor}\label{sec:publications_as_coauthor}\else
\section[Publications as coauthor]{Publications as coauthor}\label{sec:publications_as_coauthor}\fi
}
\setboolean{isCoauthorList}{true}\printbibliography[
keyword=coAuthor,
keyword=refereed,
heading=subbibliography,
title={Peer-reviewed articles},
resetnumbers=true
]
\end{refsection}}{}\ifthenelseproperty{compilation}{acknowledgement}{\vspace*{-1em}\section*{Acknowledgements}\label{sec:acknowledgement}The authors thank M.~Landreman and A.~Kaptanoglu for valuable guidance throughout this project.
This thanks extends to J.~Geiger for supplying the \gls{W7X} equilibrium and S.~Henneberg,C.~Nührenberg,B.~Jang,S.~Lazerson and I.~Ali for valuable feedback during the project.
Furthermore,we thank the reviewers for spotting several mistakes in the initial manuscript.
Last,but not least,we would like to express our sincere gratitude to the communities behind the multiple open-source software packages on which this work was built: \allowbreak \texttt{Hydra},\allowbreak \texttt{numpy},\allowbreak \texttt{jax},\allowbreak \texttt{optax},\allowbreak \texttt{jaxopt},\allowbreak \texttt{flax},\allowbreak \texttt{equinox},\allowbreak \texttt{orbax},\allowbreak \texttt{desc} and \texttt{matplotlib}.
The experiments in this publication were carried out on A100 accelerator infrastructure of the Max Planck Computing and Data Facilities cluster 'RAVEN'.
T.T.~is supported by a grant from the Simons Foundation (Grant No. 601966).
D.P.~is funded through the SciDAC program by the US Department of Energy,Office of Fusion Energy Science and Office of Advanced Scientific Computing Research under contract number DE-AC02-09CH11466,DE-SC0022005,and by the Simons Foundation/SFARI (560651).
This work was supported by a DOE Distinguished Scientist Award via DOE Contract DE-AC02-09CH11466 at the Princeton Plasma Physics Laboratory.
This work has been carried out within the framework of the EUROfusion Consortium,funded by the European Union via the Euratom Research and Training Programme (Grant Agreement No 101052200 - EUROfusion).
Views and opinions expressed are,however,those of the author(s) only and do not necessarily reflect those of the European Union or the European Commission.
Neither the European Union nor the European Commission can be held responsible for them.
}{}\ifthenelseproperty{compilation}{affidavit}{\thispagestyle{empty}\ifthenelseproperty{compilation}{clsdefineschapter}{\ifKOMA \addchap[Statutory declaration]{Statutory declaration}\else
\chapter[Statutory declaration]{Statutory declaration}\fi
}{\ifKOMA \addsec[Statutory declaration]{Statutory declaration}\else
\section[Statutory declaration]{Statutory declaration}\fi
}
I hereby declare in accordance with the examination regulations that I myself have written this document,that no other sources as those indicated were used and all direct and indirect citations are properly designated,that the document handed in was neither fully nor partly subject to another examination procedure or published and that the content of the electronic exemplar is identical to the printing copy.
\Signature{\getproperty{document}{location}}{\textsc{ \ifluatex \IfSubStr{\getproperty{author}{firstname}}{TODO}{\getproperty{author}{firstname}}{\FirstWord{\getproperty{author}{firstname}}
}
\else
\getproperty{author}{firstname}\fi
\getproperty{author}{familyname}}}
}{}\ifthenelseproperty{compilation}{bibliography}{\printbibliography
}{}\typeout{----- END OF DOCUMENT -----}\end{document}\typeout{----- END OF MAIN -----}\typeout{----- BEGIN MAIN -----}\typeout{----- BEGIN MAIN -----}